\documentclass{article}

\usepackage{microtype}
\usepackage{graphicx}
\usepackage{subcaption}
\usepackage{booktabs} 
\usepackage{bm}
\usepackage{multirow}
\usepackage{makecell}    
\usepackage{booktabs}    
\usepackage{pifont} 
\usepackage{url}
\usepackage{amsmath}
\usepackage{hyperref}

\usepackage[preprint]{icml2026}
\usepackage{amsmath}
\usepackage{amssymb}
\usepackage{mathtools}
\usepackage{amsthm}

\usepackage[capitalize,noabbrev]{cleveref}

\theoremstyle{plain}

\theoremstyle{definition}

\theoremstyle{remark}

\usepackage[textsize=tiny]{todonotes}

\icmltitlerunning{Single-Stage Signal Attenuation Diffusion Model for Low-Light Image Enhancement and Denoising}

\begin{document}

\twocolumn[
  \icmltitle{Single-Stage Signal Attenuation Diffusion Model for  \\
    Low-Light Image Enhancement and Denoising}

  \icmlsetsymbol{equal}{*}

  \begin{icmlauthorlist}
    \icmlauthor{Ying Liu}{yyy,comp}
    \icmlauthor{Junchao Zhang}{yyy,comp}
    \icmlauthor{Caiyun Wu}{yyy,comp}
  \end{icmlauthorlist}

\icmlaffiliation{yyy}{School of Aeronautics and Astronautics, Central South University, Changsha 410083, China}
\icmlaffiliation{comp}{Hunan Provincial Key Laboratory of Optic-Electronic Intelligent Measurement and Control, Changsha 410083, China}
  \icmlcorrespondingauthor{Ying Liu}{244611097@csu.edu.cn}
  \icmlcorrespondingauthor{Junchao Zhang}{junchaozhang@csu.edu.cn}
  \icmlcorrespondingauthor{Caiyun Wu}{244611076@csu.edu.cn}
]

\printAffiliationsAndNotice{}  

\begin{abstract}
  Diffusion models excel at image restoration via probabilistic modeling of forward noise addition and reverse denoising, and their ability to handle complex noise while preserving fine details makes them well-suited for Low-Light Image Enhancement (LLIE). Mainstream diffusion based LLIE methods either adopt a two-stage pipeline or an auxiliary correction network to refine U-Net outputs, which severs the intrinsic link between enhancement and denoising and leads to suboptimal performance owing to inconsistent optimization objectives. To address these issues, we propose the Signal Attenuation Diffusion Model (SADM), a novel diffusion process that integrates the signal attenuation mechanism into the diffusion pipeline, enabling simultaneous brightness adjustment and noise suppression in a single stage.
  Specifically, the signal attenuation coefficient simulates the inherent signal attenuation of low-light degradation in the forward noise addition process, encoding the physical priors of low-light degradation to explicitly guide reverse denoising toward the concurrent optimization of brightness recovery and noise suppression, thereby eliminating the need for extra correction modules or staged training relied on by existing methods. We validate that our design maintains consistency with Denoising Diffusion Implicit Models(DDIM) via multi-scale pyramid sampling, balancing interpretability, restoration quality, and computational efficiency.
  
\end{abstract}

\section{Introduction}
Low-light images, commonly encountered in critical applications such as nighttime surveillance, autonomous driving, and remote sensing, are often degraded by insufficient brightness, pronounced noise, and loss of fine details. These degradation artifacts severely hinder downstream computer vision performance, cementing LLIE as a pivotal yet challenging research topic for its joint goals of brightness enhancement, noise suppression, and detail preservation.
\begin{figure*}[ht]
	\vskip 0.2in
	\begin{center}
		\vspace*{-10pt} 
		\centering
		\includegraphics[width=0.8\linewidth]{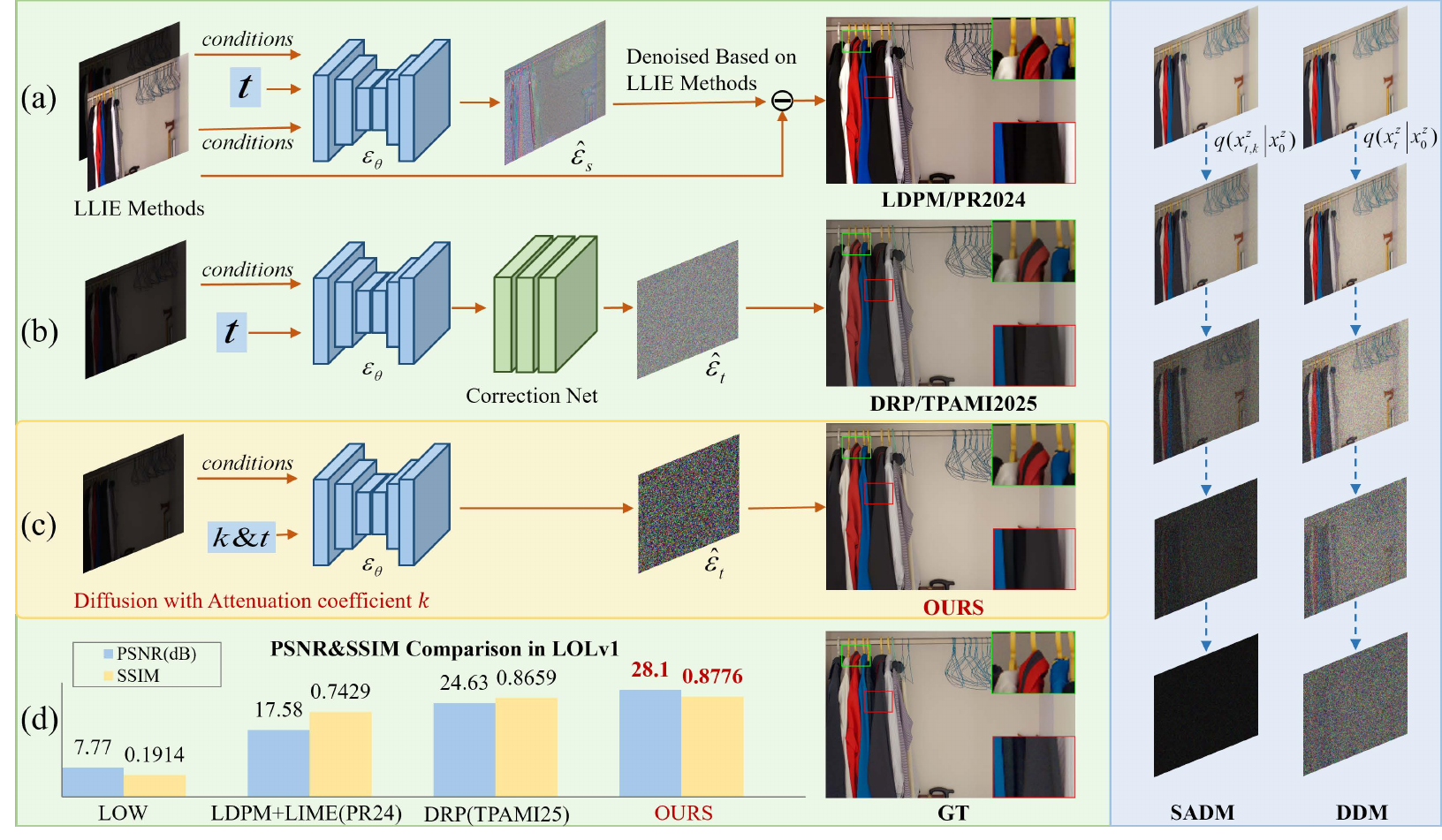}
		\caption{Comparison of Architectural Workflows and Mechanisms for Three Diffusion Model-Based Low-Light Image Enhancement and Denoising Methods. Left: Workflow comparisons of three approaches: (a) two-stage LIME+LDPM, (b) DRP with a correction network, and (c) our single-stage SADM; (d) PSNR/SSIM performance on the LOLv1 dataset, where our SADM achieves the best results. 
		Right: Comparison of forward noise addition timesteps between SADM and standard DDM, verifying the better adaptability of SADM's signal attenuation mechanism to low-light scenarios.	}
		\label{fig:teasor}
	\end{center}
\end{figure*}

Traditional LLIE methods, largely based on the Retinex theory, seek to separate an image into illumination and reflectance components. However, these approaches frequently introduce visual artifacts in noisy regions, suffer from color distortion, and struggle to balance global brightness adjustment with local detail fidelity. The advent of deep learning, particularly through Convolutional Neural Networks (CNNs) and Transformers, has led to significant advances. Nonetheless, many state-of-the-art(SOTA) methods adopt a multi-stage design, typically decoupling brightness enhancement and denoising into sequential sub-tasks. This separation not only increases model complexity but also prevents synergistic optimization, often leading to suboptimal performance and residual artifacts.

Denoising Diffusion Probabilistic Models(DDPMs) have recently emerged as a powerful paradigm for generative modeling and image restoration, distinguished by their forward noise injection and reverse denoising process. Their exceptional capability in modeling complex data distributions and recovering high-frequency details presents a promising avenue for LLIE. However, existing diffusion-based LLIE methods predominantly remain within a two-stage framework, following pipelines such as ``enhance-then-denoise" or ``diffusion-plus-correction". This decoupled strategy fails to capture the intrinsic interdependence between illumination enhancement and noise removal, while also compromising training and inference efficiency.

To overcome these limitations, we propose a novel SADM. Our core innovation is the integration of a signal attenuation mechanism into the diffusion process, thereby constructing a single-stage conditional diffusion network for LLIE. This mechanism introduces an attenuation factor during the forward process, ensuring that after $T$ diffusion steps, the image converges not only to pure noise but also to the attenuated intensity level characteristic of low-light conditions. Consequently, the reverse denoising process inherently unifies the tasks of ``brightness enhancement" and ``noise suppression" into one coherent operation. We provide rigorous mathematical derivations for the forward and reverse processes under this unified framework, enabling synergistic optimization with a single network. Furthermore, we employ DDIM sampling to enhance inference efficiency without sacrificing quality.

Figure \ref{fig:teasor} provides a comparative overview of different diffusion-based LLIE approaches. Subfigure (a) illustrates the disjoint two-stage LIME\cite{LIME} +LDPM pipeline. Subfigure (b) depicts Diff-Retinex++ (DRP), which incorporates a prior-guided refinement module within the diffusion denoiser. In contrast, our SADM (Subfigure (c)) achieves both enhancement and denoising within a single, streamlined diffusion model. Quantitative results on the LOLv1 dataset (Subfigure (d)) demonstrate that SADM achieves superior performance in terms of PSNR and SSIM. The right panel further validates the advantage of our signal attenuation mechanism in modeling low-light characteristics compared to a standard diffusion process.
Our main contributions are as follows: 
\begin{itemize}
	\item We introduce a novel SADM that unifies brightness enhancement and noise suppression within a single-stage framework. This approach derives formal diffusion formulations, incorporating a signal attenuation mechanism, to establish a solid mathematical foundation for joint optimization;
	\item We incorporate adaptive image prior guidance to mitigate color deviation and improve illumination estimation, enhancing the model's robustness across diverse scenarios;
	\item We conduct extensive experiments on multiple public datasets. Results show our method outperforms SOTA approaches on PSNR and SSIM metrics, validating its effectiveness and generalization in complex low-light scenarios.
\end{itemize}

\section{Related Works}
LLIE has seen the evolution of three primary paradigms, which sequentially emerged as research advanced: traditional prior based methods, deep learning-based methods, and diffusion model based methods.

{\bf{Traditional Methods}} are mostly based on image intensity priors or manually designed optimization models, with their core principle being to achieve enhancement through linear or nonlinear intensity transformation. Although Histogram Equalization(HE)\cite{4146204,4429280,2011Contextual} improves image brightness from the perspective of intensity distribution, it lacks targeted consideration for local detail information. While the Retinex theory \cite{557356,8304597,9056796} balances global and local properties by integrating traditional optimization methods, it is plagued by issues such as manually designed enhancement strategies, severe noise, and color deviation.
While computationally efficient and data-independent, these methods suffer from limited generalization to complex real-world scenarios due to their reliance on hand-crafted priors, often leading to over-amplification of noise, loss of detail, or unnatural color rendition.

{\bf{Deep Learning based Methods}} dominate recent research, with Convolutional Neural Networks (CNNs) \cite{9369102,Lv2018MBLLEN,2021Beyond} and Transformers \cite{wang2023ultra,Zamir2021RestormerET} as two mainstream architectures. A branch of these methods integrates Retinex theory \cite{9056796,Cai2023RetinexformerOR,9879970} and adopts a two-stage decomposition-enhancement pipeline: decomposing low-light images into illumination and reflectance maps, then refining the maps to achieve enhancement.This data-driven paradigm significantly outperforms traditional methods by learning complex nonlinear mappings from large datasets. However, these models, particularly the two-stage Retinex-based approaches, are inherently discriminative. They often suffer from several drawbacks, including uneven illumination enhancement, artifacts, color distortion, and the accumulation of pixel errors, especially when handling extreme low-light conditions or complex noise patterns not well-represented in the training data.

{\bf{Diffusion Model based Methods}} have garnered increasing attention, building on the remarkable success of DDPM \cite{DDPM} in image generation \cite{IDDPM, CDM}. Recent research has gained traction by formulating LLIE as an image generation task, leveraging the generative advantages of diffusion models to address limitations of direct enhancement methods. Unlike discriminative deep learning models that learn a direct input-output mapping, diffusion models explicitly learn the data distribution through a forward noise-adding and reverse denoising process. This fundamental shift offers superior potential for modeling the joint distribution of clean images and their low-light counterparts, enabling more natural co-optimization of brightness enhancement and noise suppression, and yielding results with higher perceptual quality. Despite continuous performance gains, most existing diffusion-based LLIE methods still adhere to the two-stage paradigm, with improvements confined to conditional guidance optimization, efficiency-performance trade-offs, and regularization strategies.

Specifically, LPDM \cite{LDPM} employs diffusion models as post-processing denoisers for base enhanced results, adhering to the ``enhancement-then-denoising" pipeline. PyDiff \cite{PyDiff} achieves a balance between efficiency and performance through pyramid diffusion and global correctors, following a logic of ``diffusion optimization-then-output correction". GSAD \cite{GSAD} adopts global structure-aware and uncertainty-guided dual regularization to enhance detail preservation. Meanwhile, AGLLDiff \cite{AGLLDiff}, LightenDiffusion \cite{LD}, AnlightenDiff \cite{AnlightenDiff}, Reti-Diff \cite{Reti-Diff}, and Diff-Retinex++ \cite{DRP} all rely on stepwise designs, including attribute extraction-then-generation, decomposition-then-enhancement, and dual-module iterative training, and they all fail to unify the enhancement and denoising tasks into a single framework.

In summary, the evolution from traditional to deep learning methods was driven by the need for greater adaptability and modeling power. The current shift towards diffusion models is motivated by their unique generative capability to handle inherent ambiguities in LLIE more effectively than discriminative models. However, current diffusion-based LLIE methods focus on conditional guidance and process optimization but lack a unified framework for brightness enhancement and noise suppression. In contrast, our proposed single-stage diffusion model integrates a signal attenuation mechanism to achieve joint modeling of the two sub-tasks, eliminating stepwise pipelines and fundamentally addressing the synergistic optimization challenge of two-stage methods.

\section{Methods}
\subsection{Signal Attenuation Diffusion Model}
We build upon the forward diffusion process of DDPM, whose formulation is given by ${x_t} = {a_t}{x_{t - 1}} + {b_t}{\varepsilon _t}$ with ${\varepsilon _t}\sim \mathcal{N}(0,\mathbf{I})$, $t = 1,2,...,T$. For LLIE adaptation, we introduce an attenuation factor ${k_t}$, leading to the forward noise addition-attenuation process:
\begin{equation}
	\label{xt_k}
	x_t = {k_t}{\rm{(}}{a_t}{x_{t - 1}} + {b_t}{\varepsilon _t}{\rm{), }}{\varepsilon _t}\sim N{\rm{(0,I)}}
\end{equation}
The ${k_t}$ attenuates signal alongside noise addition: as the image evolves to pure noise, its signal magnitude approaches zero to mimic low-light characteristics. This unifies denoising and enhancement into a single task, removing the need for separate pipelines.
By expanding ${x_t} = {k_t}{\rm{(}}{a_t}{x_{t - 1}} + {b_t}{\varepsilon _t}{\rm{)}}$ step-by-step, we obtain:
\begin{equation}
	\label{xt_expond}
	\begin{aligned}
		{x_t} &= {k_t}{\rm{(}}{a_t}{x_{t - 1}} + {b_t}{\varepsilon _t}{\rm{)}} \\
		&= {k_t}{a_t}{\rm{(}}{k_{t - 1}}{\rm{(}}{a_{t - 1}}{x_{t - 2}} + {b_{t - 1}}{\varepsilon _{t - 1}}{\rm{))}} + {k_t}{b_t}{\varepsilon _t} \\
		&= {k_t}{k_{t - 1}}{a_t}{a_{t - 1}}{\rm{(}}{k_{t - 2}}{\rm{(}}{a_{t - 2}}{x_{t - 2}} + {b_{t - 2}}{\varepsilon _{t - 2}}{\rm{))}} \\
		&\quad + {k_t}{k_{t - 1}}{a_t}{b_{t - 1}}{\varepsilon _{t - 1}} + {k_t}{b_t}{\varepsilon _t} \\
		&\;\;\vdots \\
		&= {k_t}{k_{t - 1}} \cdots {k_2}{k_1}{a_t}{a_{t - 1}} \cdots {a_2}{a_1}{x_0} \\
		&\quad + {k_t}{k_{t - 1}} \cdots {k_2}{k_1}{a_t}{a_{t - 1}} \cdots {a_2}{b_1}{\varepsilon _1} \\
		&\quad + \cdots + {k_t}{k_{t - 1}}{a_t}{b_{t - 1}}{\varepsilon _{t - 1}} + {k_t}{b_t}{\varepsilon _t}
	\end{aligned}
\end{equation}
Herein, $x_0$ denotes the original normal image; $k_t$, $a_t$, and $b_t$ are hyperparameters that control the noise scheduling process of the diffusion model. Specifically, $k_t$ represents the signal attenuation coefficient with a fixed step-wise attenuation rate of 0.999(i.e.,$k_t = 0.999 \times  k_{t-1},t>0$). $a_t$ denotes the attenuation coefficient and $b_t$ refers to the noise coefficient with a value in the interval $[0,1]$. In addition, $\varepsilon_t$ denotes random noise sampled from a standard Gaussian distribution. 

To derive the equivalent variance, we superpose multiple independent Gaussian distributions (i.e., ${\varepsilon _t}$). The derivation process is as follows:
\begin{equation}
	\label{crump_k}
	\begin{array}{l}
		\mathop \prod \limits_{s = 1}^t k_s^2\mathop \prod \limits_{s = 1}^t a_s^2 + \mathop \prod \limits_{s = 1}^t k_s^2\mathop \prod \limits_{s = 2}^t a_s^2b_1^2 +  \cdots  \\
		 + \mathop \prod \limits_{s = t - 1}^t k_s^2\mathop \prod \limits_{s = t}^t a_s^2b_{t - 1}^2 + k_t^2b_t^2\\
		= \mathop \prod \limits_{s = 1}^t k_s^2{\rm{(}}\mathop \prod \limits_{s = 1}^t a_s^2 + \mathop \prod \limits_{s = 2}^t a_s^2b_1^2{\rm{)}} +  \cdots  \\
		+ \mathop \prod \limits_{s = t - 1}^t k_s^2\mathop \prod \limits_{s = t}^t a_s^2b_{t - 1}^2 + k_t^2b_t^2\\
		= \mathop \prod \limits_{s = 2}^t k_s^2\mathop \prod \limits_{s = 3}^t a_s^2{\rm{(}}k_{\rm{1}}^2a_{\rm{2}}^2{\rm{(}}a_1^2 + b_1^2{\rm{) + }}b_2^2{\rm{)}} + \cdots  \\
		  + \mathop \prod \limits_{s = t - 1}^t k_s^2\mathop \prod \limits_{s = t}^t a_s^2b_{t - 1}^2 + k_t^2b_t^2
	\end{array}
\end{equation}
By letting $k_{t - 1}^4a_t^2 + b_t^2 = k_t^2$ with $k_0 = 1$ (i.e., $a_1^2 + b_1^2 = k_1^2$), we simplify the above expression to obtain:
\begin{equation}
	\label{kt4}
	\begin{array}{l}
	\mathop \prod \limits_{s = 1}^t k_s^2\mathop \prod \limits_{s = 1}^t a_s^2 + \mathop \prod \limits_{s = 1}^t k_s^2\mathop \prod \limits_{s = 2}^t a_s^2b_1^2 \\ 
	+ \cdots 	+ \mathop \prod \limits_{s = t - 1}^t k_s^2\mathop \prod \limits_{s = t}^t a_s^2b_{t - 1}^2 + k_t^2b_t^2{\rm{ = }}k_t^4
	\end{array}
\end{equation}
Thus, the equivalent variance of these independent Gaussian distributions is $k_t^4 - \mathop \prod \limits_{s = 1}^t k_s^2\mathop \prod \limits_{s = 1}^t a_s^2$. 

To simplify notation, we let $\mathop \prod \limits_{s = 1}^t k_s = \bar{k}_t$ and $\mathop \prod \limits_{s = 1}^t a_s = \bar{a}_t$, leading to the following forward relational expressions:
\begin{equation}
	\label{x0xt}
	\begin{array}{*{20}{l}}
		{{x_t} = {k_t}{a_t}{x_{t - 1}} + {k_t}{b_t}{\varepsilon _t}}&{\left( {\rm{a}} \right)}\\
		{{x_t} = {{\bar k}_t}{{\bar a}_t}{x_0} + \sqrt {k_t^4 - \bar k_t^2\bar a_t^2} {{\tilde \varepsilon }_t}}&{\left( {\rm{b}} \right)}\\
		{{x_0} = \frac{1}{{{{\bar k}_t}{{\bar a}_t}}}{\rm{(}}{x_t} - \sqrt {k_t^4 - \bar k_t^2\bar a_t^2} {{\tilde \varepsilon }_t}{\rm{)}}}&{\left( {\rm{c}} \right)}
	\end{array}
\end{equation}
Here, $x_0$ denotes the original normal image; ${\varepsilon _t}$ and ${\tilde \varepsilon _t}$ are independent random noise samples following the standard Gaussian distribution. Specifically, sub-formula (a) describes the step-wise noise injection process that transforms the image from step $t-1$ ($x_{t-1}$) to step $t$ ($x_t$). Sub-formula (b) represents the direct noise injection process across $t$ steps, mapping the original image $x_0$ to the noisy image $x_t$. Sub-formula (c) corresponds to the reverse restoration process, which recovers the original image $x_0$ from the noisy image $x_t$. Furthermore, ${\tilde \varepsilon _t}$ denotes a standard Gaussian distribution distinct from ${\varepsilon _t}$; leveraging the Markov chain property, we derive the following formulation:
\begin{equation}
	\label{q_t2_t1}
	q{\rm{(}}{{\mathop{\rm x}\nolimits} _{t - 1}}{\rm{|}}{{\mathop{\rm x}\nolimits} _t}{\rm{) = }}\frac{{q{\rm{(}}{{\mathop{\rm x}\nolimits} _t}{\rm{|}}{{\mathop{\rm x}\nolimits} _{t - 1}}{\rm{)}}q{\rm{(}}{{\mathop{\rm x}\nolimits} _{t - 1}}{\rm{|}}{{\mathop{\rm x}\nolimits} _0}{\rm{)}}}}{{q{\rm{(}}{{\mathop{\rm x}\nolimits} _t}{\rm{|}}{{\mathop{\rm x}\nolimits} _0}{\rm{)}}}} \propto N\left( {{x_{t - 1}}{\rm{; }}{\mu _q}{\rm{,}}\sigma _q^2{\rm{I}}} \right)
\end{equation}
\begin{equation}
	\label{q_combined}
	\begin{aligned}
		{\mu _q} &= \frac{{{a_t}\left( {k_{t - 1}^4 - \bar k_{t - 1}^2\bar a_{t - 1}^2} \right)}}{{{k_t}\left( {k_t^2 - \bar k_{t - 1}^2\bar a_t^2} \right)}}{x_t} + \frac{{b_t^2{\rm{ }}{{\bar k}_{t - 1}}{{\bar a}_{t - 1}}}}{{\left( {k_t^2 - \bar k_{t - 1}^2\bar a_t^2} \right)}}{x_0} \\
		\sigma _q^2 &= \frac{{b_t^2\left( {k_{t - 1}^4 - \bar k_{t - 1}^2\bar a_{t - 1}^2} \right)}}{{k_t^2 - \bar k_{t - 1}^2\bar a_t^2}}
	\end{aligned}
\end{equation}
 By substituting Eq.(\ref{x0xt}-c) into the expression for ${\mu _q}$, we derive: 
\begin{equation}
	\label{uq}
	{\mu _q} = \frac{1}{{{k_t}{a_t}}}{x_t} - \frac{{b_t^2{\rm{ }}}}{{{a_t}\sqrt {k_t^2 - \bar k_{t - 1}^2\bar a_t^2} }}{\tilde \varepsilon _t}
\end{equation}
We set $\sigma _\theta ^2 = \sigma _q^2$ and define ${\mu _\theta } = \frac{1}{{{k_t}{a_t}}}{x_t} - \frac{{b_t^2}}{{\sqrt {\frac{{k_t^2}}{{a_t^2}} - \overline k _{t - 1}^2\overline a _{t - 1}^2}  \cdot a_t^2}}{f_\theta }\left( {{x_t},t} \right)$. With these definitions, the final optimization objective can be transformed into: 
\begin{equation}
	\label{klqp}
	\begin{array}{l}
		\mathop {\arg \min }\limits_\theta  {D_{KL}}\left( {q\left( {{x_{t - 1}}|{x_t},{x_0}} \right)\left\| {{p_\theta }\left( {{x_{t - 1}}|{x_t}} \right)} \right.} \right)\\
		= \mathop {\arg \min }\limits_\theta  \frac{1}{{2\sigma _q^2\left( t \right)}}\frac{{{{\left( {b_t^2} \right)}^2}}}{{\left( {\frac{{k_t^2}}{{a_t^2}} - \overline k _{t - 1}^2\overline a _{t - 1}^2} \right) \cdot a_t^4}}\left[ {\left\| {{f_\theta }\left( {{x_t},t} \right) - {{\tilde \varepsilon }_t}} \right\|_2^2} \right]
	\end{array}
\end{equation}

\subsection{DDIM Sampling}
Stepwise iterative computation with DDPM typically demands considerable computational resources. To mitigate this issue, we integrate the multi-scale pyramid strategy and DDIM sampling to improve image processing efficiency. Multi-scale pyramid sampling reduces computational load effectively by lowering image resolution, while DDIM optimizes the sampling logic within the diffusion process—significantly boosting inference speed without sacrificing generation quality. The combination of these two techniques enables the construction of an efficient, high-fidelity generation framework.

DDIM relaxes the constraints of the traditional Markov chain via mathematical derivation. It eliminates the requirement for retraining DDPM, only necessitating modifications to the sampler. These modifications result in a significant increase in sampling speed, while maintaining the fidelity of the generated images.

We assume $p\left( {{x_{t - 1}}|{x_t},{x_0}} \right)$ follows ${p_\theta }\left( {{x_{t - 1}}|{x_t},{x_0}} \right) = \mathcal{N}\left( {{x_{t - 1}};m{x_0} + n{x_t},{\sigma ^2}} \right)$, which differs from the Markov chain process in DDPM. From the forward noise addition formula derived earlier, we have: ${x_t} = {\bar k_t}{\bar a_t}{x_0} + \sqrt {k_t^4 - \bar k_t^2\bar a_t^2} {\tilde \varepsilon _t}$. Substituting this into ${x_{t - 1}} = m{x_0} + n{x_t} + \sigma \varepsilon $ yields:
\begin{equation}
	\label{ddim_1}
	{x_{t - 1}} = m{x_0} + n{\bar k_t}{\bar a_t}{x_0} + n\sqrt {k_t^4 - \bar k_t^2\bar a_t^2} {\tilde \varepsilon _t} + \sigma \varepsilon 
\end{equation}
Here, ${\tilde \varepsilon _t}$ , ${\tilde \varepsilon _{t - 1}}$ , and ${\varepsilon}$ are mutually independent and all follow the standard Gaussian distribution. 

Similarly, we can derive the expression for ${x_{t - 1}}$ using the forward noise addition formula: ${x_{t - 1}} = {\bar k_{t - 1}}{\bar a_{t - 1}}{x_0} + \sqrt {k_{t - 1}^4 - \bar k_{t - 1}^2\bar a_{t - 1}^2} {\tilde \varepsilon _{t - 1}}$. By combining like terms, we enforce consistency of the corresponding coefficients, leading to:
\begin{equation}
	\label{ddim_mn}
	\begin{aligned}
		m + n\bar{k}_t \bar{a}_t &= \bar{k}_{t-1} \bar{a}_{t-1} \\
		n^2 (k_t^4 - \bar{k}_t^2 \bar{a}_t^2) + \sigma^2 &= k_{t-1}^4 - \bar{k}_{t-1}^2 \bar{a}_{t-1}^2
	\end{aligned}
\end{equation}
Upon resolving this system of equations, we deduce:
\begin{equation}
	\label{ddim_3}
	\begin{array}{l}
		m = {{\bar k}_{t - 1}}{{\bar a}_{t - 1}} - {{\bar k}_t}{{\bar a}_t}\frac{{\sqrt {k_{t - 1}^4 - \bar k_{t - 1}^2\bar a_{t - 1}^2 - {\sigma ^2}} }}{{\sqrt {k_t^4 - \bar k_t^2\bar a_t^2} }}\\
		n = \frac{{\sqrt {k_{t - 1}^4 - \bar k_{t - 1}^2\bar a_{t - 1}^2 - {\sigma ^2}} }}{{\sqrt {k_t^4 - \bar k_t^2\bar a_t^2} }}
	\end{array}
\end{equation}
Substituting these derived values of $m$ and $n$ back into the expression ${x_{t - 1}} = m{x_0} + n{x_t} + \sigma \varepsilon$, we arrive at:
\begin{equation}
	\label{ddim_4}
	\begin{aligned}
		x_{t-1} &= \left( \bar{k}_{t-1} \bar{a}_{t-1} - \bar{k}_t \bar{a}_t 
		\frac{\sqrt{k_{t-1}^4 - \bar{k}_{t-1}^2 \bar{a}_{t-1}^2 - \sigma^2}}
		{\sqrt{k_t^4 - \bar{k}_t^2 \bar{a}_t^2}} \right) x_0 \\
		&+ \frac{\sqrt{k_{t-1}^4 - \bar{k}_{t-1}^2 \bar{a}_{t-1}^2 - \sigma^2}}
		{\sqrt{k_t^4 - \bar{k}_t^2 \bar{a}_t^2}} x_t + \sigma \varepsilon
	\end{aligned}
\end{equation}
Following the same principle as reverse denoising, we substitute ${x_0}$ into the aforementioned formula (Eq.(\ref{ddim_4})), yielding the final derived form as follows:
\begin{equation}
	\label{DDIM_xt_1}
	\begin{alignedat}{2}
		&x_{t-1} &&= \frac{\bar{k}_{t-1} \bar{a}_{t-1}}{\bar{k}_t \bar{a}_t} \bigl(x_t - \sqrt{k_t^4 - \bar{k}_t^2 \bar{a}_t^2} \,\tilde{\varepsilon}_t \bigr) \\
		&&&+ \sqrt{k_{t-1}^4 - \bar{k}_{t-1}^2 \bar{a}_{t-1}^2 - \sigma^2} \,\tilde{\varepsilon}_t + \sigma \varepsilon
	\end{alignedat}
\end{equation}
Throughout this derivation, we do not invoke the Markov chain process. Consequently, the ${x_{t - 1}}$ in Eq.(\ref{DDIM_xt_1}) can be generalized to any prior step ${x_p}$:
\begin{equation}
	\label{xp}
	{x_p} = \frac{{{{\bar k}_p}{{\bar a}_p}}}{{{{\bar k}_t}{{\bar a}_t}}}({x_t} - \sqrt {k_t^4 - \bar k_t^2\bar a_t^2} {\tilde \varepsilon _t}) + \sqrt {k_p^4 - \bar k_p^2\bar a_p^2 - {\sigma ^2}} {\tilde \varepsilon _t} + \sigma \varepsilon 
\end{equation}
To mitigate perturbations caused by randomness, the variance parameter $\sigma$ is set to 0 in our model configuration.
\subsection{Network Architecture and Loss Function}
\begin{figure*}[ht]
	\vskip 0.2in
	\begin{center}
	\vspace*{-10pt} 
		\centering
		\includegraphics[width=\linewidth]{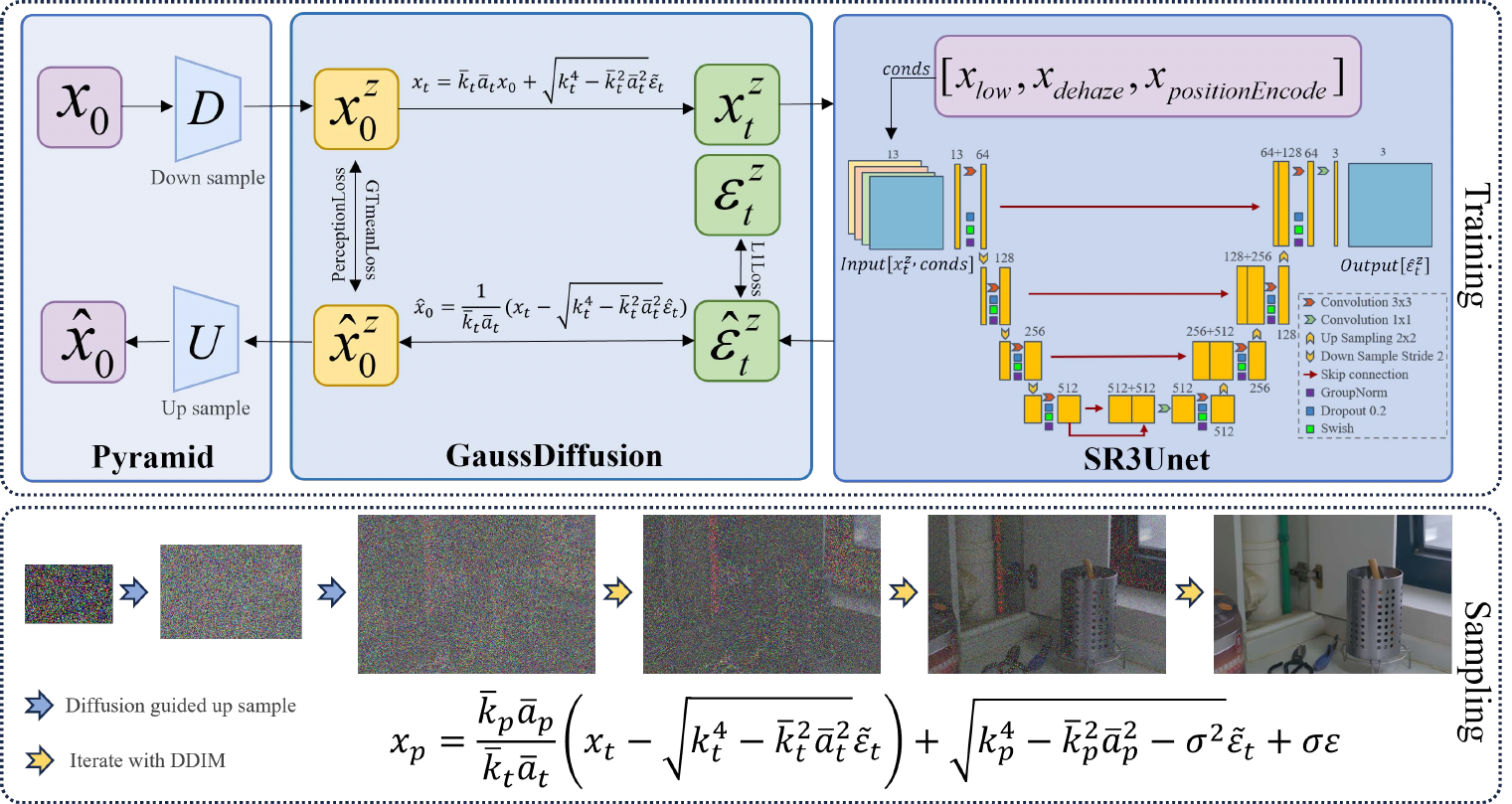}
		\caption{Framework of SADM. Under the multi-scale pyramid framework, the normally exposed image $x_0$ is downsampled to generate $x_0^z$, which is then contaminated with Gaussian noise to produce the noisy image $x_t^z$. The low-light image $x_{\text{low}}$, the image prior condition $x_{\text{dehaze}}$ (the detailed design is presented in Appendix B.5) and the position encoding $x_{\text{positionEncode}}$ (following the setting in PyDiff\cite{PyDiff}) are taken as joint conditional information, which is fed into the SR3Unet alongside the noisy image $x_t^z$. Distinct loss functions are leveraged to constrain both the noise predicted by the SR3Unet and the derived target image. Ultimately, DDIM is employed to conduct sampling and inference based on random Gaussian noise.	}
		\label{fig:framework}
	\end{center}
\end{figure*}
This section details the design principles underlying the proposed method, focusing on the network architecture and loss function, with the overall framework depicted in Figure \ref{fig:framework}. The workflow of the method is divided into two core stages, namely training and sampling. Their specific designs are detailed as follows:

{\bf{Training Stage}}: For an input image, a preset image scale is adaptively selected based on the current training time step. After processing by the Gaussian diffusion model, a noisy image at the corresponding scale is generated. Subsequently, the noisy image, current time step information, and image conditions are jointly fed into the SR3Unet \cite{SR3} network, which generates the predicted noise. The training process is constrained by a dual loss function. On one hand, the L1 loss function is used to calculate the error between the predicted noise and the ground-truth noise generated by the Gaussian diffusion model. On the other hand, the Gaussian diffusion model restores the initial image ${\hat x_0}$ from the noisy image. The loss between ${\hat x_0}$ and the original input image ${x_0}$ is computed via GTmeanLoss and PerceptionLoss to enhance restoration accuracy.

\textbf{\bfseries Sampling Stage}: First, a noise map sampled from a standard Gaussian distribution is generated. Then, leveraging the pyramid structure, the DDIM sampling algorithm is applied to iteratively restore the image. To improve cross-scale restoration quality, a differentiated iteration strategy is adopted during sampling. For intra-scale iteration (i.e., within the same scale), the image is updated directly by invoking the DDIM prediction formula. For cross-scale iteration (i.e., during scale switching), the upsampled image is generated by applying diffusion-based noise addition logic. This approach preserves more detailed information and outperforms traditional interpolation-based upsampling. The total number of iteration steps for the entire sampling process is 10, with the step allocation across different scales as follows: [1, 1, 1, 2, 2, 2, 4, 4, 4, 4].

{\bf{Loss Funcion}}:
Drawing on the brightness alignment loss function proposed by Liao et al. \cite{GTMean}, this method addresses the brightness mismatch between the enhanced image and Ground Truth (GT) in low-light image enhancement by dynamically balancing the original loss term and the brightness alignment loss term. Specifically, the brightness alignment loss term scales the brightness of the enhanced image to match that of GT using a scaling factor ${\lambda _{GT}} = \frac{{{\rm E}[y]}}{{{\rm E}[f(x)]}}$, where ${\rm E}[ \cdot ]$ denotes the average brightness of the image. The weight $W$ is calculated by quantifying the brightness distribution difference between the two via Bhattacharyya Distance, ranging from 0 to 1. In this study, the original loss is set as L1 Loss with a weight of 1.0, which imposes constraints on ${\hat x_0}$.
\begin{equation}
	\label{LossGT}
	\begin{aligned}
		L_{\text{GT}}(f(x),y) &= W \cdot  \underbrace{ L_{L1}(f(x),y)}_{\text{original loss}} \\
		&\quad + (1 - W) \cdot \underbrace{L_{L1}\left( \lambda_{\text{GT}}f(x),y \right)}_{\text{brightness-aligned loss}}
	\end{aligned}
\end{equation}
To ensure perceptual-level visual consistency, we leverage the feature responses of the VGG-19 network and adjust weights for specific layers to ensure perceptual-level visual consistency. The loss adopts the $L1$ criterion to measure feature differences, with a weight of 0.01.
\begin{equation}
	\label{LossP}
	\begin{aligned}
		L_P &= \lambda_P \cdot [\omega_{\text{conv3\_4}} \cdot L_{L1}(\phi_{\text{conv3\_4}}(\hat{x}_0^z), \phi_{\text{conv3\_4}}(x_0^z)) \\
		&\quad + \omega_{\text{conv4\_4}} \cdot L_{L1}(\phi_{\text{conv4\_4}}(\hat{x}_0^z), \phi_{\text{conv4\_4}}(x_0^z))]
	\end{aligned}
\end{equation}
Accordingly, the final total loss function of the proposed model is formulated as:
\begin{equation}
	\label{deqn_ex1}
	{L_{Total}} = {L_{GT}}(\hat x_0^z,x_0^z) + {L_P}(\hat x_0^z,x_0^z) + {L_{L1}}({\rm{ }}\hat \varepsilon _t^z,\varepsilon _t^z\;)
\end{equation}
\section{Experiments}
\subsection{Datasets and Evaluation Metrics}
This section introduces the datasets and evaluation metrics to validate the effectiveness and generalization of the proposed SADM method. Comprehensive experiments are conducted to quantitatively and qualitatively evaluate the model’s performance against SOTA approaches.

To rigorously validate SADM, extensive comparative experiments are conducted on three widely used low-light image enhancement benchmarks: LOLv1 \cite{LOLv1}, LOLv2-Real\cite{LOLv2}, and LOLv2-Syn \cite{LOLv2}. To further evaluate the model’s generalization ability, we additionally collect images from other public low-light datasets for testing, including DICM \cite{DICM}, NPE\cite{NPE}, MEF\cite{MEF} and others.

We adopt two standard metrics in the field for quantitative evaluation: Peak Signal-to-Noise Ratio (PSNR) and Structural Similarity Index (SSIM) \cite{SSIM}, where higher values indicate better performance.
\subsection{Comparison with LLIE Methods}
\begin{figure*}[h!]
	\vskip 0.2in
	\begin{center}
		\vspace*{-10pt} 
		\centering
		\includegraphics[width=\linewidth,trim=0.0cm 0.0cm 0.0cm 0.0cm, clip]{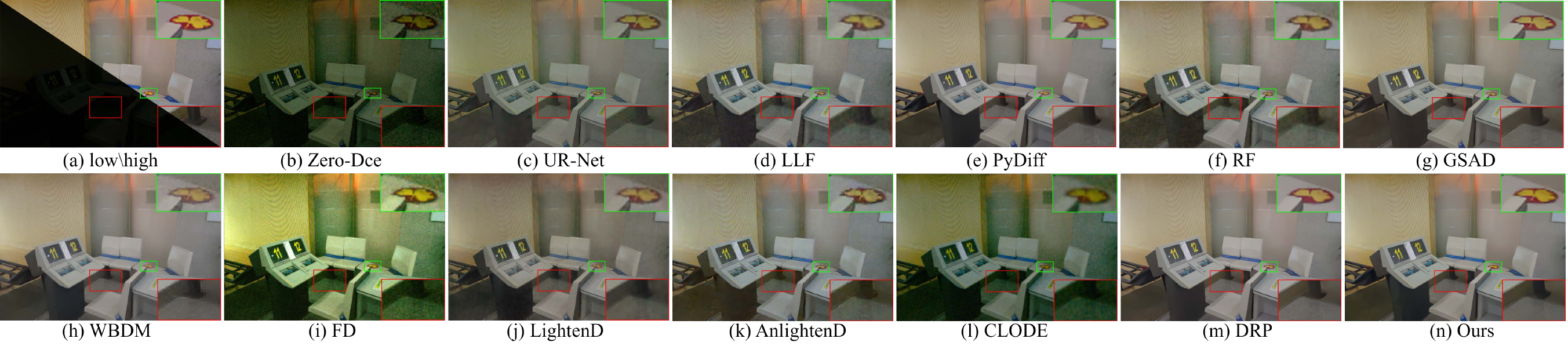}
		\caption{Visual effect comparison of different algorithms on the LOLv1 test dataset:(a)low\textbackslash high; (b)Zero-Dce; (c)UR-Net; (d)LLF; (e)PyDiff; (f)RF; (g)GSAD; (h)WBDM; (i)FD; (j)LightenD; (k)AnlightenD; (l)CLODE; (m)DRP; (n)Ours. Local magnified views are presented in the red and green boxes.}
		\label{fig:665}
	\end{center}
\end{figure*}

\begin{table*}[htbp]
	\vspace{-10pt} 
	\centering
	\fontsize{6.5pt}{8pt}\selectfont 
	\setlength{\tabcolsep}{1.5pt} 
	\caption{Quantitative evaluation results in LOLv1, LOLv2\_real, LOLv2\_syn}
	\begin{tabular}{l*{14}{c}}  
		\toprule
		Datasets & Metrics &
		\makecell*[c]{Zero-DCE\\(CVPR20)} &  
		\makecell*[c]{UR-Net\\(CVPR22)} & 
		\makecell*[c]{LLFormer\\(AAAI23)} & 
		\makecell*[c]{PyDiff\\(ICCAI23)} & 
		\makecell*[c]{RF\\(ICCV23)} & 
		\makecell*[c]{GSAD\\(NeurIPS23)} & 
		\makecell*[c]{WBDM\\(TOG23)} &  
		\makecell*[c]{FourierD\\(CVPR24)} & 
		\makecell*[c]{LightenD\\(ECCV24)} & 
		\makecell*[c]{AnlightenD\\(TIP24)} & 
		\makecell*[c]{CLODE\\(ICLR25)} &  
		\makecell*[c]{DRP\\(TPAMI25)} & 
		{OURS} \\
		\midrule
		\multirow{2}{*}{LOLv1} &{PSNR$\uparrow$}
		& 14.8607 & 19.8416 & 23.6491 & 27.0043 & 25.1526 & 27.6405 & 26.3204 & 17.5569 & 19.9187 & 20.0503  & 19.6031 &  24.6320 & \textcolor{red}{\textbf{28.0985}}
	 \\
		&{SSIM$\uparrow$} 
		& 0.5624 & 0.8243 & 0.8163 & 0.8747 & 0.8434 & 0.8752 & 0.8445 & 0.6116 & 0.8091 & 0.7498  & 0.7175 & 0.8659 & \textcolor{red}{\textbf{0.8770}}\\
		
		\midrule
		\multirow{2}{*}{LOLv2\_real} 
		&{PSNR$\uparrow$} 
		& 18.0588 & 21.0933 & 27.7489 & - & 22.7938 & 28.6497 & 28.8706 & 16.8634 & 22.6514 & 20.4606 & 17.8711 &23.2284 & \textcolor{red}{\textbf{28.9338}} \\
		& {SSIM$\uparrow$}
		& 0.5795  & 0.8576  & 0.8602  & - & 0.8387  & \textcolor{red}{\textbf{0.8930}}   & 0.8759  & 0.6075  & 0.8512  & 0.7576   & 0.6830   &0.8714 & 0.8797 \\
		\midrule
		\multirow{2}{*}{LOLv2\_syn} 
		& {PSNR$\uparrow$}
		&17.7564 & 18.2487 & 17.1632 & - & 25.6697 & 28.3251 & \textcolor{red}{\textbf{29.4624}} & 14.0556 & 21.6341 & 17.0581 & 17.2125 & 26.1412 & 26.7042 \\
		& {SSIM$\uparrow$} 
		&0.8140   & 0.8209  & 0.7842  & - & 0.9282  & 0.9426  & 0.9106  & 0.6429  & 0.8664  & 0.7066  & 0.7852 & \textcolor{red}{\textbf{0.9444}}  & 0.9224  \\
		\midrule
		\multirow{2}{*}{Efficiency} &{Time(s)/Steps}
		& 0.0024/1 & 3.1092/1 & 3.5889/1
		 & 2.4000/4 & 0.6239/1
		 & 4.9064/20  & 1.2467/10
		  & 124.26/100
		   & 1.1227/20 & 150.93/100  & 4.982/auto &  5.7947/5 & 4.0067/10	\\
		&{Param(M)} 
		& 0.0794 & 0.6470 & 24.5490 & 97.9156 & 1.6057 & 17.4402 & 22.0789 & 552.8141 & 34.9023 & 94.9960 & 0.6897  & 62.8236 & 97.8235 \\
		\bottomrule
	\end{tabular}
	\raggedright
	\vspace{5pt}
	{\scriptsize The indicator in bold \textcolor{red}{\textbf{red}} represents the best, and all algorithms' efficiency parameters were conducted on an NVIDIA GeForce RTX 4060 with the LOLv1 test dataset.}
	\label{tab:Metrics}
\end{table*}
To comprehensively evaluate the performance of the proposed SADM for LLIE, we select 12 representative  SOTA algorithms covering curve estimation-based, Retinex theory-based, diffusion model-based, and frequency domain transformation-based paradigms in LLIE for comparative experiments, including Zero-DCE\cite{Zero-DCE}, URetinexNet (UR-Net)\cite{URetinexNet}, LLFormer (LLF)\cite{LLFormer}, PyDiff\cite{PyDiff}, Retinexformer (RF)\cite{Retinexformer}, GSAD\cite{GSAD}, Wavelet-Based Diffusion Models (WBDM)\cite{WBDM}, FourierDiff (FD)\cite{Fourier}, LightenDiffusion (LightenD)\cite{LD}, AnlightenDiff (AnlightenD)\cite{AnlightenDiff}, CLODE\cite{CLODE}, and Diff-Retinex-Plus (DRP)\cite{DRP}. 

Quantitative evaluations on LOLv1, LOLv2\_real, and LOLv2\_syn (Table \ref{tab:Metrics}) demonstrate our SADM’s superior performance: we achieve the highest PSNR (28.0987) and SSIM (0.8770) on LOLv1, outperforming all SOTA methods; on LOLv2\_real, our PSNR (28.9342) ranks first, with SSIM (0.8797) only slightly lower than GSAD (0.8930); even on LOLv2\_syn where DRP leads in SSIM, our SADM still delivers competitive PSNR (26.7045) and SSIM (0.9224) values. 

\begin{figure*}[h]
	\vskip 0.2in
	\begin{center}
		\vspace*{-15pt} 
		\centering
		\includegraphics[width=\linewidth]{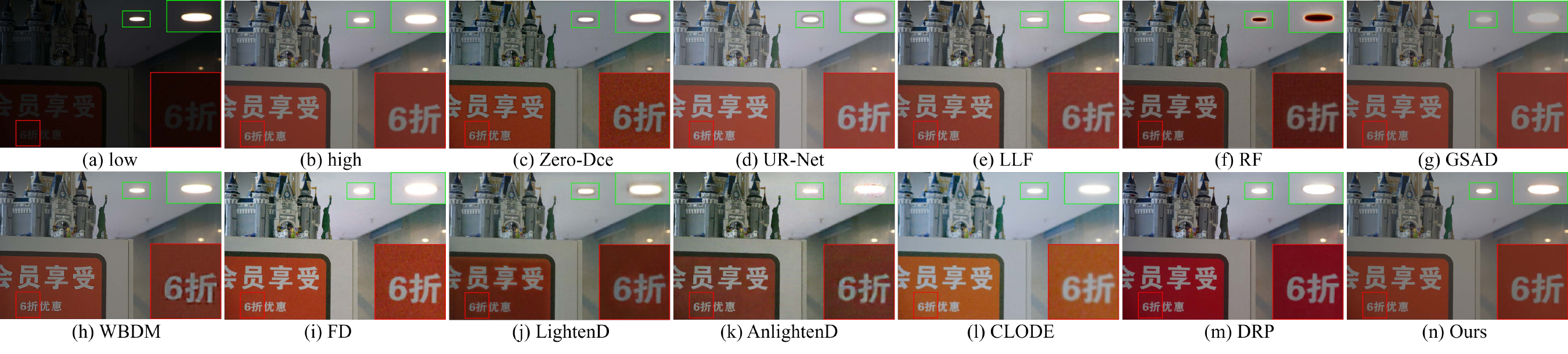}
		\caption{Visual effect comparison of different algorithms on the LOLv2\_real dataset:(a)low; (b)high; (c)Zero-Dce; (d)UR-Net; (e)LLF; (f)RF; (g)GSAD; (h)WBDM; (i)FD; (j)LightenD; (k)AnlightenD; (l)CLODE; (m)DRP; (n)Ours. Local magnified views are presented in the red and green boxes.}
		\label{fig:12_v2}
	\end{center}
\end{figure*}

\begin{figure}[t]
	\vskip 0.2in
	\begin{center}
		\centering
		\vspace*{-10pt} 
		\includegraphics[width=\linewidth,trim=0.0cm 0.0cm 0.0cm 0.0cm, clip]{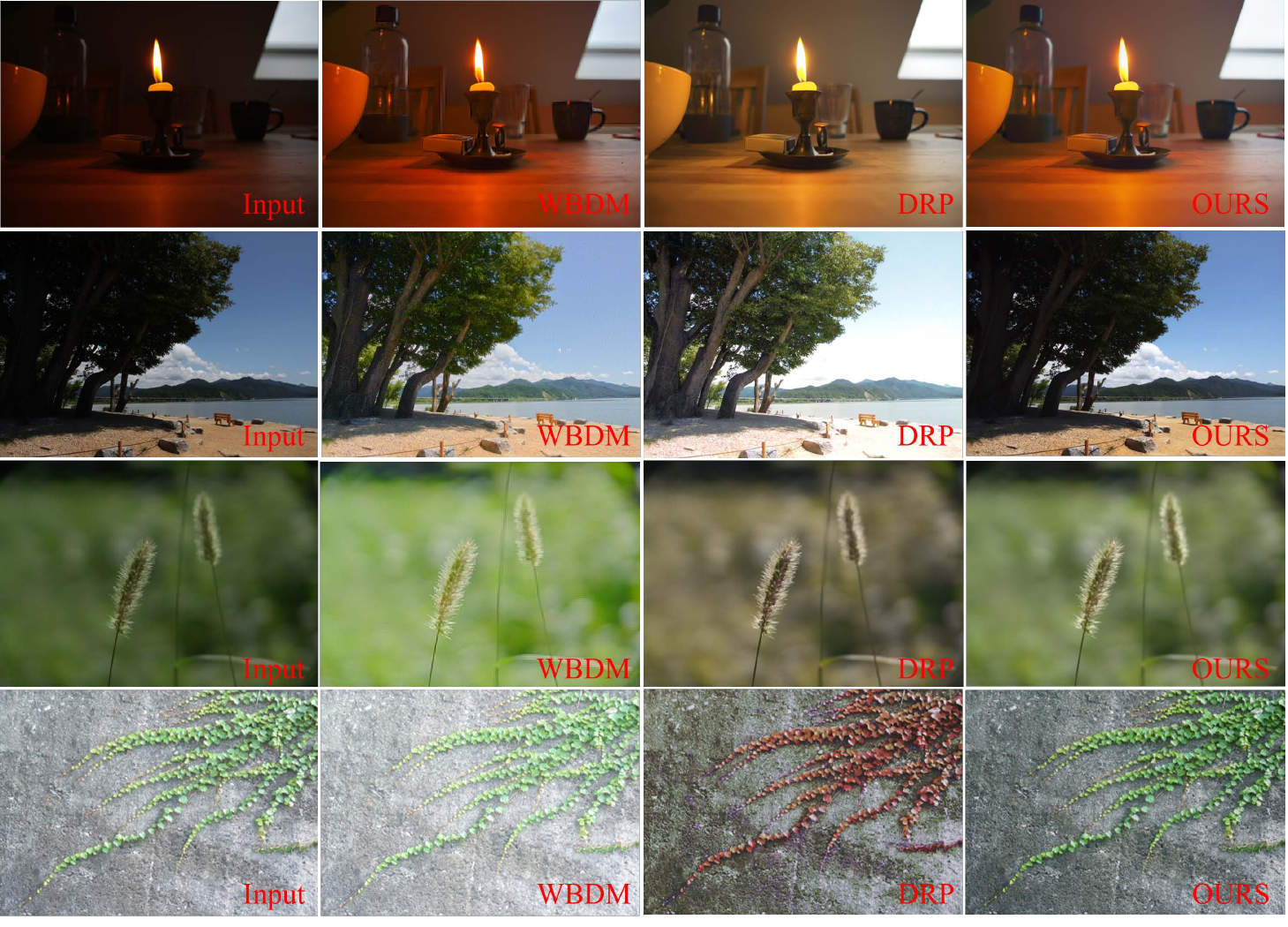}
		\caption{Performance of our SADM on unpaired (reference-free) low-light datasets (weights pre-trained on LOLv1). This figure presents visual enhancement results for input, WBDM, DRP and SADM to validate the model’s generalization ability, with a key strength in adaptive overexposure suppression.	}
		\label{fig:unpaired}
	\end{center}
\end{figure}

Figures \ref{fig:665}, \ref{fig:12_v2}, and \ref{fig:unpaired} present qualitative results of our SADM across different datasets to validate its performance and generalization, consistent with the quantitative superiority shown in Table \ref{tab:Metrics}. 

Figure \ref{fig:665} on the LOLv1 test set displays enhancement results of LLIE algorithms, aligning with our top-ranked PSNR/SSIM on this dataset. Subfigures (a) correspond to the original low-light image and GT, while (b)-(n) present outputs of competing algorithms and SADM. Red boxes focus on dark, high-noise regions and demonstrate SADM’s superior brightness recovery, noise elimination and smearing suppression. In red boxes, SADM achieves less noise than GT while preserving image details well. Green boxes highlight the golden leaf on a red background and reveal competitors’ flaws, such as color distortion, residual noise and smearing. An example is the lost petiole detail in Subfigure (m).

Figure \ref{fig:12_v2} on the LOLv2\_real test set further confirms SADM’s superiority. Green boxes target local light source areas, where SADM performs excellently. SADM adjusts local brightness effectively to achieve uniform distribution without overexposure or underexposure, while preserving surrounding textures well. This outperforms competitors that often suffer from local light distortion or blurred details. 

For generalization validation, we evaluate SADM on unpaired datasets, with results in Fig. \ref{fig:unpaired} comparing input, WBDM, DRP, and SADM(all using weights pre-trained on LOLv1) . Our method’s key strength lies in adaptive overexposure suppression; it restores the warm, natural tone of low-light indoor candle scenes without over-amplification, and preserves outdoor sky clarity by suppressing overexposure while eliminating noise. In contrast, WBDM tends to over-enhance or over-smooth textures, while DRP introduces residual noise or color distortion. These results confirm SADM’s robust generalization to unpaired real-world low-light scenes. More experimental results of our algorithm are presented in Appendix C.2.

\subsection{Ablation Study}
\begin{table}[h]
	\centering
	\vspace*{-10pt} 
	\caption{Ablation Study Results with Component Configuration}
	\label{tab:ablation_study_components}
	\resizebox{\linewidth}{!}{
		\begin{tabular}{cccc|cc}
			\toprule
			{Groups} & $\bm{k}$ & {${x_0}$ loss} & {Prior Condition} & {PSNR} & {SSIM}  \\
			\midrule
			Ab.1  &  \ding{55} & GTmean\&Percep &  dehaze & 20.9171 & 0.4547  \\
			Ab.2 & \ding{51} & GTmean  &dehaze & 27.8724 & \textcolor{red}{\textbf{0.8801}}  \\
			Ab.3 & \ding{51}  & Percep & dehaze & 27.7354 & 0.8719  \\
			Ab.4  & \ding{51} & GTmean\&Percep  & hiseq  & 26.1812 & 0.8602  \\
			Ab.5 & \ding{51}   & GTmean\&Percep &\ding{55}  & 26.6877  & 0.8620  \\
			OURS  & \ding{51}  & GTmean\&Percep & dehaze & \textcolor{red}{\textbf{28.0987}} & 0.8770  \\
			\bottomrule
		\end{tabular}}
		\vspace*{-10pt} 
\end{table}

\begin{figure}[h!]
	\vskip 0.2in
	\begin{center}
		\vspace{-10pt} 
		\centering
		\includegraphics[width=\linewidth]{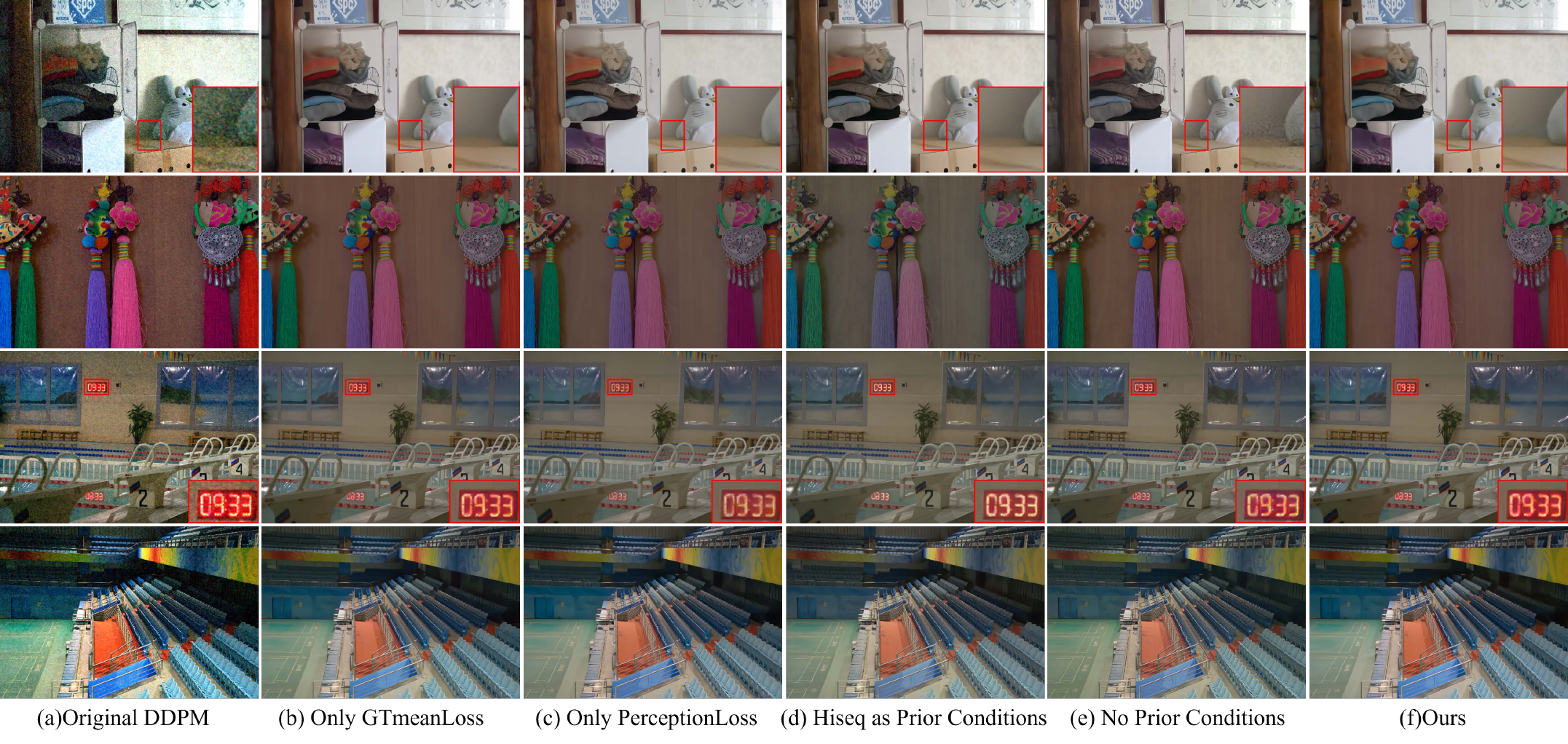}
		\caption{	Visual Effect Comparison of Ablation Study on the LOLv1test Dataset. (a)Ablation 1 original DDPM (without $k$); (b)Ablation 2 only with GTmeanLoss; (c)Ablation 3 only with PerceptionLoss; (d) Ablation 4 use histogram equalization as prior condition; (e)Ablation 5 no prior condition. Local magnified views highlight the performance differences of each ablation variant.}
		\label{fig:Ab}
	\end{center}
\end{figure}
To validate the effectiveness of the proposed modules, we conducted five ablation experiments with the following configurations: Ab.1: original DDPM(w/o $k$); Ab.2: employing only GTmeanLoss; Ab.3: employing only PerceptionLoss; Ab.4: using histogram equalization for the prior condition input; Ab.5: removing the prior condition input.

Ablation results (Figure \ref{fig:Ab}, Table \ref{tab:ablation_study_components}) show parameter $k$ significantly impacts model performance. Adding PerceptionLoss during training slightly reduces generated images’ SSIM but compensates for GTmeanLoss’s pixel-level constraint. GTmeanLoss tends to cause inconsistencies between local details and overall style, while PerceptionLoss enforces perceptual constraints to enhance visual coherence and align with human perception, while boosting PSNR. For prior condition optimization, histogram equalization on low-light images enhances dark-region details to aid feature learning but risks color distortion. A lightweight Dehaze algorithm effectively reveals dark-region features and mitigates color deviation, improving input prior effectiveness.
\section{Conclusion}
In this paper, we propose a signal attenuation diffusion model to address the limitations of the two-step ``enhancement followed by denoising" method in low-light image enhancement, which integrates brightness enhancement and noise suppression into a single task to simplify the model structure and improve processing efficiency; a key innovation is the introduction of an attenuation factor in the forward noising process, which effectively suppresses signal intensity while gradually increasing noise, making the degraded images closer to real low-light conditions. Extensive experiments on multiple public datasets and ablation studies confirm that the proposed method surpasses SOTA approaches in the key quantitative metrics PSNR and SSIM, fully demonstrating its efficacy and superiority in complex low-light scenarios.
\clearpage
\section*{Software and Data}
If our paper is accepted, we will make the source code and experimental data of the proposed SADM method publicly available in a dedicated GitHub repository, in line with the conference’s encouragement for open sharing of software and data.

\section*{Impact Statement}
Our work focuses on low-light image enhancement and denoising via diffusion models, a technology with broad practical applications in computer vision. Positively, the proposed SADM model can improve the reliability of downstream tasks (e.g., object detection, surveillance, autonomous driving) in low-light scenarios, contributing to safer and more efficient real-world systems. It also provides a more interpretable and efficient paradigm for diffusion-based image restoration, advancing fundamental research in this field.

Our work uses only publicly available, ethically sourced datasets, ensuring no privacy violations or biases are introduced during model training. There are many potential societal consequences of our work, none which we feel must be specifically highlighted here.

\bibliography{example_paper}

\begin{thebibliography}{40}
\providecommand{\natexlab}[1]{#1}
\providecommand{\url}[1]{\texttt{#1}}
\expandafter\ifx\csname urlstyle\endcsname\relax
  \providecommand{\doi}[1]{doi: #1}\else
  \providecommand{\doi}{doi: \begingroup \urlstyle{rm}\Url}\fi

\bibitem[Abdullah-Al-Wadud et~al.(2007)Abdullah-Al-Wadud, Kabir, Dewan, and
  Chae]{4146204}
Abdullah-Al-Wadud, M., Kabir, M.~H., Dewan, M. A.~A., and Chae, O.
\newblock A dynamic histogram equalization for image contrast enhancement.
\newblock In \emph{2007 Digest of Technical Papers International Conference on
  Consumer Electronics}, pp.\  1--2, 2007.
\newblock \doi{10.1109/ICCE.2007.341567}.

\bibitem[Cai et~al.(2023{\natexlab{a}})Cai, Bian, Lin, Wang, Timofte, and
  Zhang]{Cai2023RetinexformerOR}
Cai, Y., Bian, H., Lin, J., Wang, H., Timofte, R., and Zhang, Y.
\newblock Retinexformer: One-stage retinex-based transformer for low-light
  image enhancement.
\newblock \emph{2023 IEEE/CVF International Conference on Computer Vision
  (ICCV)}, pp.\  12470--12479, 2023{\natexlab{a}}.
\newblock URL \url{https://api.semanticscholar.org/CorpusID:257496232}.

\bibitem[Cai et~al.(2023{\natexlab{b}})Cai, Bian, Lin, Wang, Timofte, and
  Zhang]{Retinexformer}
Cai, Y., Bian, H., Lin, J., Wang, H., Timofte, R., and Zhang, Y.
\newblock Retinexformer: One-stage retinex-based transformer for low-light
  image enhancement.
\newblock In \emph{Proceedings of the IEEE/CVF International Conference on
  Computer Vision (ICCV)}, pp.\  12504--12513, October 2023{\natexlab{b}}.

\bibitem[Celik \& Tjahjadi(2011)Celik and Tjahjadi]{2011Contextual}
Celik, T. and Tjahjadi, T.
\newblock Contextual and variational contrast enhancement.
\newblock \emph{IEEE Transactions on Image Processing}, 20\penalty0
  (12):\penalty0 3431--3441, 2011.

\bibitem[Chan et~al.(2024)Chan, Siu, Chan, and Anthony~Chan]{AnlightenDiff}
Chan, C.-Y., Siu, W.-C., Chan, Y.-H., and Anthony~Chan, H.
\newblock Anlightendiff: Anchoring diffusion probabilistic model on low light
  image enhancement.
\newblock \emph{IEEE Transactions on Image Processing}, 33:\penalty0
  6324--6339, 2024.
\newblock \doi{10.1109/TIP.2024.3486610}.

\bibitem[Feifan~Lv \& Lim(2018)Feifan~Lv and Lim]{Lv2018MBLLEN}
Feifan~Lv, Feng~Lu, J.~W. and Lim, C.
\newblock Mbllen: Low-light image/video enhancement using cnns.
\newblock \emph{British Machine Vision Conference}, 2018.

\bibitem[Guo et~al.(2020)Guo, Li, Guo, Loy, Hou, Kwong, and Cong]{Zero-DCE}
Guo, C.~G., Li, C., Guo, J., Loy, C.~C., Hou, J., Kwong, S., and Cong, R.
\newblock Zero-reference deep curve estimation for low-light image enhancement.
\newblock In \emph{Proceedings of the IEEE conference on computer vision and
  pattern recognition (CVPR)}, pp.\  1780--1789, June 2020.

\bibitem[Guo et~al.(2017)Guo, Li, and Ling]{LIME}
Guo, X., Li, Y., and Ling, H.
\newblock Lime: Low-light image enhancement via illumination map estimation.
\newblock \emph{IEEE Transactions on Image Processing}, 26\penalty0
  (2):\penalty0 982--993, 2017.
\newblock \doi{10.1109/TIP.2016.2639450}.

\bibitem[He et~al.(2025)He, Fang, Zhang, Tang, Huang, Li, guo, Li, and
  Farsiu]{Reti-Diff}
He, C., Fang, C., Zhang, Y., Tang, L., Huang, J., Li, K., guo, z., Li, X., and
  Farsiu, S.
\newblock Reti-diff: Illumination degradation image restoration with
  retinex-based latent diffusion model.
\newblock In Yue, Y., Garg, A., Peng, N., Sha, F., and Yu, R. (eds.),
  \emph{International Conference on Representation Learning}, volume 2025, pp.\
   43332--43352, 2025.

\bibitem[Ho et~al.(2020)Ho, Jain, and Abbeel]{DDPM}
Ho, J., Jain, A., and Abbeel, P.
\newblock Denoising diffusion probabilistic models.
\newblock In Larochelle, H., Ranzato, M., Hadsell, R., Balcan, M., and Lin, H.
  (eds.), \emph{Advances in Neural Information Processing Systems}, volume~33,
  pp.\  6840--6851. Curran Associates, Inc., 2020.

\bibitem[HOU et~al.(2023)HOU, Zhu, Hou, LIU, Zeng, and Yuan]{GSAD}
HOU, J., Zhu, Z., Hou, J., LIU, H., Zeng, H., and Yuan, H.
\newblock Global structure-aware diffusion process for low-light image
  enhancement.
\newblock In \emph{Thirty-seventh Conference on Neural Information Processing
  Systems}, 2023.
\newblock URL \url{https://openreview.net/forum?id=bv9mmH0LGF}.

\bibitem[Ibrahim \& Pik~Kong(2007)Ibrahim and Pik~Kong]{4429280}
Ibrahim, H. and Pik~Kong, N.~S.
\newblock Brightness preserving dynamic histogram equalization for image
  contrast enhancement.
\newblock \emph{IEEE Transactions on Consumer Electronics}, 53\penalty0
  (4):\penalty0 1752--1758, 2007.
\newblock \doi{10.1109/TCE.2007.4429280}.

\bibitem[Jiang et~al.(2023)Jiang, Luo, Fan, Han, and Liu]{WBDM}
Jiang, H., Luo, A., Fan, H., Han, S., and Liu, S.
\newblock Low-light image enhancement with wavelet-based diffusion models.
\newblock \emph{ACM Transactions on Graphics (TOG)}, 42\penalty0 (6):\penalty0
  1--14, 2023.

\bibitem[Jiang et~al.(2024)Jiang, Luo, Liu, Han, and Liu]{LD}
Jiang, H., Luo, A., Liu, X., Han, S., and Liu, S.
\newblock Lightendiffusion: Unsupervised low-light image enhancement with
  latent-retinex diffusion models.
\newblock In \emph{European Conference on Computer Vision}, 2024.

\bibitem[Jobson et~al.(1997)Jobson, Rahman, and Woodell]{557356}
Jobson, D., Rahman, Z., and Woodell, G.
\newblock Properties and performance of a center/surround retinex.
\newblock \emph{IEEE Transactions on Image Processing}, 6\penalty0
  (3):\penalty0 451--462, 1997.
\newblock \doi{10.1109/83.557356}.

\bibitem[Jung et~al.(2025)Jung, Kim, and Kim]{CLODE}
Jung, D., Kim, D., and Kim, T.~H.
\newblock Continuous exposure learning for low-light image enhancement using
  neural odes.
\newblock In \emph{International Conference on Learning Representations}, 2025.
\newblock URL \url{https://api.semanticscholar.org/CorpusID:278498134}.

\bibitem[Lee et~al.(2012)Lee, Lee, and Kim]{DICM}
Lee, C., Lee, C., and Kim, C.-S.
\newblock Contrast enhancement based on layered difference representation.
\newblock In \emph{2012 19th IEEE International Conference on Image
  Processing}, pp.\  965--968, 2012.
\newblock \doi{10.1109/ICIP.2012.6467022}.

\bibitem[Li et~al.(2022)Li, Guo, and Loy]{9369102}
Li, C., Guo, C., and Loy, C.~C.
\newblock Learning to enhance low-light image via zero-reference deep curve
  estimation.
\newblock \emph{IEEE Transactions on Pattern Analysis and Machine
  Intelligence}, 44\penalty0 (8):\penalty0 4225--4238, 2022.
\newblock \doi{10.1109/TPAMI.2021.3063604}.

\bibitem[Li et~al.(2018)Li, Liu, Yang, Sun, and Guo]{8304597}
Li, M., Liu, J., Yang, W., Sun, X., and Guo, Z.
\newblock Structure-revealing low-light image enhancement via robust retinex
  model.
\newblock \emph{IEEE Transactions on Image Processing}, 27\penalty0
  (6):\penalty0 2828--2841, 2018.
\newblock \doi{10.1109/TIP.2018.2810539}.

\bibitem[Liao et~al.(2025)Liao, Hao, Hong, and Wang]{GTMean}
Liao, J., Hao, S., Hong, R., and Wang, M.
\newblock Gt-mean loss: A simple yet effective solution for brightness mismatch
  in low-light image enhancement.
\newblock In \emph{Proceedings of the IEEE/CVF International Conference on
  Computer Vision (ICCV)}, pp.\  6112--6121, October 2025.

\bibitem[Lin et~al.(2024)Lin, Ye, Chen, Fu, Wang, Chai, Xing, Zhu, and
  Ding]{AGLLDiff}
Lin, Y., Ye, T., Chen, S., Fu, Z., Wang, Y., Chai, W., Xing, Z., Zhu, L., and
  Ding, X.
\newblock Aglldiff: Guiding diffusion models towards unsupervised training-free
  real-world low-light image enhancement, 2024.

\bibitem[Lv et~al.(2024)Lv, Zhang, Wang, Zheng, Zhong, Li, and Nie]{Fourier}
Lv, X., Zhang, S., Wang, C., Zheng, Y., Zhong, B., Li, C., and Nie, L.
\newblock Fourier priors-guided diffusion for zero-shot joint low-light
  enhancement and deblurring.
\newblock In \emph{Proceedings of the IEEE/CVF Conference on Computer Vision
  and Pattern Recognition}, pp.\  25378--25388, 2024.

\bibitem[Ma et~al.(2015)Ma, Zeng, and Wang]{MEF}
Ma, K., Zeng, K., and Wang, Z.
\newblock Perceptual quality assessment for multi-exposure image fusion.
\newblock \emph{IEEE Transactions on Image Processing}, 24\penalty0
  (11):\penalty0 3345--3356, 2015.
\newblock \doi{10.1109/TIP.2015.2442920}.

\bibitem[Nichol \& Dhariwal(2021)Nichol and Dhariwal]{IDDPM}
Nichol, A.~Q. and Dhariwal, P.
\newblock Improved denoising diffusion probabilistic models.
\newblock In Meila, M. and Zhang, T. (eds.), \emph{Proceedings of the 38th
  International Conference on Machine Learning}, volume 139 of
  \emph{Proceedings of Machine Learning Research}, pp.\  8162--8171. PMLR,
  18--24 Jul 2021.
\newblock URL \url{https://proceedings.mlr.press/v139/nichol21a.html}.

\bibitem[Panagiotou \& Bosman(2024)Panagiotou and Bosman]{LDPM}
Panagiotou, S. and Bosman, A.~S.
\newblock Denoising diffusion post-processing for low-light image enhancement.
\newblock \emph{Pattern Recognition}, 156:\penalty0 110799, 2024.
\newblock ISSN 0031-3203.
\newblock \doi{https://doi.org/10.1016/j.patcog.2024.110799}.
\newblock URL
  \url{https://www.sciencedirect.com/science/article/pii/S0031320324005508}.

\bibitem[Ren et~al.(2020)Ren, Yang, Cheng, and Liu]{9056796}
Ren, X., Yang, W., Cheng, W.-H., and Liu, J.
\newblock Lr3m: Robust low-light enhancement via low-rank regularized retinex
  model.
\newblock \emph{IEEE Transactions on Image Processing}, 29:\penalty0
  5862--5876, 2020.
\newblock \doi{10.1109/TIP.2020.2984098}.

\bibitem[Saharia et~al.(2023)Saharia, Ho, Chan, Salimans, Fleet, and
  Norouzi]{SR3}
Saharia, C., Ho, J., Chan, W., Salimans, T., Fleet, D.~J., and Norouzi, M.
\newblock Image super-resolution via iterative refinement.
\newblock \emph{IEEE Transactions on Pattern Analysis and Machine
  Intelligence}, 45\penalty0 (4):\penalty0 4713--4726, 2023.
\newblock \doi{10.1109/TPAMI.2022.3204461}.

\bibitem[Wang et~al.(2013)Wang, Zheng, Hu, and Li]{NPE}
Wang, S., Zheng, J., Hu, H.-M., and Li, B.
\newblock Naturalness preserved enhancement algorithm for non-uniform
  illumination images.
\newblock \emph{IEEE Transactions on Image Processing}, 22\penalty0
  (9):\penalty0 3538--3548, 2013.
\newblock \doi{10.1109/TIP.2013.2261309}.

\bibitem[Wang et~al.(2023{\natexlab{a}})Wang, Zhang, Shen, Luo, Stenger, and
  Lu]{LLFormer}
Wang, T., Zhang, K., Shen, T., Luo, W., Stenger, B., and Lu, T.
\newblock Ultra-high-definition low-light image enhancement: A benchmark and
  transformer-based method.
\newblock In \emph{Proceedings of the AAAI Conference on Artificial
  Intelligence}, volume~37, pp.\  2654--2662, 2023{\natexlab{a}}.

\bibitem[Wang et~al.(2023{\natexlab{b}})Wang, Zhang, Shen, Luo, Stenger, and
  Lu]{wang2023ultra}
Wang, T., Zhang, K., Shen, T., Luo, W., Stenger, B., and Lu, T.
\newblock Ultra-high-definition low-light image enhancement: A benchmark and
  transformer-based method.
\newblock In \emph{Proceedings of the AAAI Conference on Artificial
  Intelligence}, volume~37, pp.\  2654--2662, 2023{\natexlab{b}}.

\bibitem[Wang et~al.(2004)Wang, Bovik, Sheikh, and Simoncelli]{SSIM}
Wang, Z., Bovik, A., Sheikh, H., and Simoncelli, E.
\newblock Image quality assessment: from error visibility to structural
  similarity.
\newblock \emph{IEEE Transactions on Image Processing}, 13\penalty0
  (4):\penalty0 600--612, 2004.
\newblock \doi{10.1109/TIP.2003.819861}.

\bibitem[Wei et~al.(2018)Wei, Wang, Yang, and Liu]{LOLv1}
Wei, C., Wang, W., Yang, W., and Liu, J.
\newblock Deep retinex decomposition for low-light enhancement.
\newblock In \emph{British Machine Vision Conference}, 2018.

\bibitem[Wu et~al.(2022{\natexlab{a}})Wu, Weng, Zhang, Wang, Yang, and
  Jiang]{9879970}
Wu, W., Weng, J., Zhang, P., Wang, X., Yang, W., and Jiang, J.
\newblock Uretinex-net: Retinex-based deep unfolding network for low-light
  image enhancement.
\newblock In \emph{2022 IEEE/CVF Conference on Computer Vision and Pattern
  Recognition (CVPR)}, pp.\  5891--5900, 2022{\natexlab{a}}.
\newblock \doi{10.1109/CVPR52688.2022.00581}.

\bibitem[Wu et~al.(2022{\natexlab{b}})Wu, Weng, Zhang, Wang, Yang, and
  Jiang]{URetinexNet}
Wu, W., Weng, J., Zhang, P., Wang, X., Yang, W., and Jiang, J.
\newblock Uretinex-net: Retinex-based deep unfolding network for low-light
  image enhancement.
\newblock In \emph{2022 IEEE/CVF Conference on Computer Vision and Pattern
  Recognition (CVPR)}, pp.\  5891--5900, 2022{\natexlab{b}}.
\newblock \doi{10.1109/CVPR52688.2022.00581}.

\bibitem[Yang et~al.(2021)Yang, Wang, Huang, Wang, and Liu]{LOLv2}
Yang, W., Wang, W., Huang, H., Wang, S., and Liu, J.
\newblock Sparse gradient regularized deep retinex network for robust low-light
  image enhancement.
\newblock \emph{IEEE Transactions on Image Processing}, 30:\penalty0
  2072--2086, 2021.
\newblock \doi{10.1109/TIP.2021.3050850}.

\bibitem[Yi et~al.(2025)Yi, Xu, Zhang, Tang, and Ma]{DRP}
Yi, X., Xu, H., Zhang, H., Tang, L., and Ma, J.
\newblock Diff-retinex++: Retinex-driven reinforced diffusion model for
  low-light image enhancement.
\newblock \emph{IEEE Transactions on Pattern Analysis and Machine
  Intelligence}, 47\penalty0 (8):\penalty0 6823--6841, 2025.
\newblock \doi{10.1109/TPAMI.2025.3563612}.

\bibitem[Zamir et~al.(2021)Zamir, Arora, Khan, Hayat, Khan, and
  Yang]{Zamir2021RestormerET}
Zamir, S.~W., Arora, A., Khan, S.~H., Hayat, M., Khan, F.~S., and Yang, M.-H.
\newblock Restormer: Efficient transformer for high-resolution image
  restoration.
\newblock \emph{2022 IEEE/CVF Conference on Computer Vision and Pattern
  Recognition (CVPR)}, pp.\  5718--5729, 2021.
\newblock URL \url{https://api.semanticscholar.org/CorpusID:244346144}.

\bibitem[Zhang et~al.(2021)Zhang, Guo, Ma, Liu, and Zhang]{2021Beyond}
Zhang, Y., Guo, X., Ma, J., Liu, W., and Zhang, J.
\newblock Beyond brightening low-light images.
\newblock \emph{International Journal of Computer Vision}, 129\penalty0 (2),
  2021.

\bibitem[Zhou et~al.(2023)Zhou, Yang, and Yang]{PyDiff}
Zhou, D., Yang, Z., and Yang, Y.
\newblock Pyramid diffusion models for low-light image enhancement.
\newblock In Elkind, E. (ed.), \emph{Proceedings of the Thirty-Second
  International Joint Conference on Artificial Intelligence, {IJCAI-23}}, pp.\
  1795--1803. International Joint Conferences on Artificial Intelligence
  Organization, 8 2023.
\newblock \doi{10.24963/ijcai.2023/199}.
\newblock URL \url{https://doi.org/10.24963/ijcai.2023/199}.
\newblock Main Track.

\bibitem[Zhu et~al.(2023)Zhu, Li, Wang, He, and Yao]{CDM}
Zhu, Y., Li, Z., Wang, T., He, M., and Yao, C.
\newblock Conditional text image generation with diffusion models.
\newblock \emph{2023 IEEE/CVF Conference on Computer Vision and Pattern
  Recognition (CVPR)}, pp.\  14235--14244, 2023.
\newblock URL \url{https://api.semanticscholar.org/CorpusID:259203172}.

\end{thebibliography}
\bibliographystyle{icml2026}

\newpage
\appendix
\onecolumn

\section{Detailed Related Works}

Traditional LLIE methods are mostly based on image intensity priors or manually designed optimization models, with their core principle being to achieve enhancement through linear or nonlinear intensity transformation. Although Histogram Equalization (HE \cite{4146204,4429280,2011Contextual}) improves image brightness from the perspective of intensity distribution, it lacks targeted consideration for local detail information. While the Retinex theory \cite{557356,8304597,9056796} balances global and local properties by integrating traditional optimization methods, it is plagued by issues such as manually designed enhancement strategies, severe noise, and color deviation.

Deep learning-based LLIE methods adopt data-driven learning strategies to learn the mapping from low-light images to normally exposed images, and have emerged as a key research focus in recent years. The two typical network architectures for this task are convolutional neural networks (CNNs)\cite{9369102,Lv2018MBLLEN,2021Beyond} and Transformers \cite{wang2023ultra,Zamir2021RestormerET}.

Furthermore, several deep learning studies have incorporated the physics-inspired Retinex theory \cite{9056796,Cai2023RetinexformerOR,9879970}. These deep learning-based Retinex methods primarily adopt a two-stage framework consisting of decomposition and enhancement. They use convolutional neural networks to decompose an image into an illumination map and a reflectance map. Subsequently, component restoration is achieved by adjusting these two maps, thereby realizing image enhancement. However, two-stage Retinex methods suffer from several drawbacks, including uneven illumination enhancement, artifacts, color distortion, and the accumulation of pixel errors.

Researchers have achieved remarkable success in applying DDPM \cite{DDPM} to the field of image generation \cite{IDDPM,CDM}. An approach that has garnered significant attention in recent research involves treating low-light image enhancement as an image generation task, and leveraging the generative advantages of diffusion models to address the limitations of direct image enhancement methods.

While diffusion model-based low-light image enhancement algorithms have achieved continuous performance breakthroughs, the core design of most methods still revolves around stepwise processing and fails to break free from the inherent two-stage framework. Their key improvements mainly focus on the optimization of conditional diffusion strategies, the balance between efficiency and performance, and regularization-based enhancement, with specific manifestations as follows:

LPDM \cite{LDPM} adopts a post-processing paradigm, treating the diffusion model as a noise suppression module. It first obtains a base enhanced result using traditional enhancement models and then performs diffusion-based denoising, which is essentially a lightweight extension of the ``enhancement first, denoising later" two-stage framework; 
PyDiff \cite{PyDiff} achieves a balance between efficiency and performance through pyramid diffusion and a global corrector, following the core logic of ``optimizing the diffusion process first, then correcting the output";
GSAD\cite{GSAD} employs a dual regularization strategy of global structure awareness and uncertainty guidance, generating pixel-level uncertainty maps via a pre-trained uncertainty estimation model to enhance the detail preservation capability and robustness of the diffusion model;  
AGLLDiff \cite{AGLLDiff} innovates the enhancement logic with an ``attribute-guided" strategy but still needs to first decompose low-light images using a pre-trained RNet to extract reflectance maps, then guide the color optimization of diffusion sampling based on these maps, relying on the stepwise control logic of ``attribute extraction first, generation guidance later"; 
LightenDiffusion \cite{LD} represents a typical stepwise design, explicitly splitting the task into two independent networks for ``decomposition and enhancement": it first derives the reflectance map and illumination map through Retinex decomposition, then trains a diffusion model based on these decomposition results to achieve enhancement, relying on stepwise modeling to ensure stability; 
AnlightenDiff \cite{AnlightenDiff} focuses on the diffusion anchoring mechanism, with its training process divided into two steps—separately training a central encoder and then training a noise predictor—reducing artifacts through stepwise optimization and essentially adhering to a stepwise iterative design; 
Reti-Diff\cite{Reti-Diff} adopts a dual-module architecture of ``prior generation and feature enhancement": it first extracts Retinex priors via the RPE module for preliminary reconstruction, then generates high-quality priors through dual-branch diffusion training, depending on a stepwise workflow to ensure information consistency; 
Diff-Retinex++ \cite{DRP} integrates the Denoising Diffusion Model with the Retinex-driven Mixture of Experts model, adopting a stepwise fusion approach of ``diffusion-based generation plus Retinex physical constraint correction".

In summary, while these methods differ in their specific approaches, they all revolve around two core aspects of diffusion models: conditional guidance and process optimization. Whether by decomposing the task, imposing attribute constraints, or innovating mechanisms, most algorithms fail to treat enhancement and denoising as a unified task. In contrast, the single-stage diffusion model we propose achieves unified modeling of brightness enhancement and noise suppression by integrating the signal attenuation mechanism into the diffusion modeling process. Without the need to decompose the task or design multi-stage workflows, our model not only simplifies the network structure and training logic while improving efficiency but also fundamentally addresses the synergistic optimization challenge caused by stepwise processing in two-stage methods.

\section{Detailed Methods}
\subsection{Basic Diffusion Model}
Diffusion models are generative models composed of two processes: the forward process and the reverse process. The forward process, also known as the diffusion process, and the reverse process are both parameterized Markov chains. In the forward process, given an original image, Gaussian noise is incrementally added over T steps, ultimately yielding a completely random noise image. The reverse process, by contrast, recovers the original image from the noise image.
The core of a diffusion model resides in the process of noise prediction and denoising. As Michelangelo famously stated, \emph{The statue is already in the marble; I only remove the unnecessary parts}. Analogously, the diffusion process entails gradually denoising the noise image until the original image is fully recovered.

\subsubsection{Forward Noise Addition Process}
Let ${x_0}$ represent the original image; the subsequent process ${x_{t - 1}} \to {x_t}$ corresponds to a continuous noise addition procedure. The relationship between two adjacent variables is linear and can be modeled as:

\begin{equation}
	\label{deqn_ex1}
	{x_t} = {a_t}{x_{t - 1}} + {b_t}{\varepsilon _t}{\rm{, }}{\varepsilon _t}\sim N{\rm{(0,I)}}
\end{equation}

Here, ${\varepsilon _t}$ denotes noise that follows a standard Gaussian distribution. Since ${x_{t - 1}}$ contains more information,  ${a_t}$ is an attenuation coefficient ranging between 0 and 1. Similarly, the noise coefficient ${b_t}$ follows an increasing trend, with values also between 0 and 1. Let $a_t^2 + b_t^2 = 1$; and ${\overline a _t} = {a_t}{a_{t - 1}} \cdots {a_2}{a_1}$. From the superposition property of independent normal distributions, we can derive:

\begin{equation}
	\label{deqn_ex1}
	{x_t} = {\overline a _t}{x_0} + \sqrt {\left( {1 - \overline a _t^2} \right)} {\overline \varepsilon  _t},{\overline \varepsilon  _t} \sim N{\rm{(0,I)}}
\end{equation}

where ${\overline \varepsilon  _t}$ and ${\varepsilon _t}$ are mutually independent, and both follow the standard Gaussian distribution.
The linear process above can also be regarded as sampling from a Gaussian distribution, and its specific implementation relies on the reparameterization trick. The complete forward transition process is written as follows:

\begin{equation}
	\label{deqn_ex1}
	{x_t} \sim q\left( {{x_t}|{x_{t - 1}}} \right) = N\left( {{x_t}{\rm{;}}{a_t}{x_{t - 1}}{\rm{,}}\left( {1 - a_t^2} \right){\rm{I}}} \right)
\end{equation}

\begin{equation}
	\label{deqn_ex1}
	{x_t} \sim q\left( {{x_t}|{x_0}} \right) = N\left( {{x_t}{\rm{;}}{{\overline a }_t}{x_0}{\rm{,}}\left( {1 - \overline a _t^2} \right){\rm{I}}} \right)
\end{equation}

\subsubsection{Reverse Denoising Process}
The reverse process aims to recover the original data from Gaussian noise. Since the noise added in each step of the forward process is minimal, we assume that $p\left( {{x_{t - 1}}|{x_t}} \right)$ also follows a Gaussian distribution. Define $p$ as a parameterized Gaussian distribution to approximate the reverse diffusion process, where $\theta $ denotes the parameters of the neural network. The reverse process is also a Markov chain process, which can be expressed as:

\begin{equation}
	\label{deqn_ex1}
	{p_\theta }\left( {{x_{t - 1}}|{x_t}} \right) = N\left( {{x_{t - 1}}{\rm{;}}{\mu _\theta }\left( {{x_t},t} \right){\rm{,}}\sigma _\theta ^2\left( {{x_t},t} \right)} \right)
\end{equation}

Here, ${\mu _\theta }$ and $\sigma _\theta ^{\rm{2}}$ are the mean and variance to be estimated.

From the forward process and Bayes' theorem, we have:

\begin{equation}
	\label{deqn_ex1}
	q\left( {{x_{t - 1}}|{x_t},{x_0}} \right) = \frac{{q\left( {{x_t}|{x_{t - 1}},{x_0}} \right)q\left( {{x_{t - 1}}|{x_0}} \right)}}{{q\left( {{x_t}|{x_0}} \right)}}
\end{equation}

Substituting the relational expression derived from the forward process and simplifying it, we obtain:

\begin{equation}
	\label{deqn_ex1}
	\begin{array}{c}
		q\left( {{x_{t - 1}}|{x_t},{x_0}} \right) = \frac{{q\left( {{x_t}|{x_{t - 1}},{x_0}} \right)q\left( {{x_{t - 1}}|{x_0}} \right)}}{{q\left( {{x_t}|{x_0}} \right)}} = \frac{{N\left( {{x_t}{\rm{;}}{a_t}{x_{t - 1}}{\rm{,}}\left( {1 - a_t^2} \right){\rm{I}}} \right)N\left( {{x_{t - 1}}{\rm{;}}{{\overline a }_{t - 1}}{x_0}{\rm{,}}\left( {1 - \overline a _{t - 1}^2} \right){\rm{I}}} \right)}}{{N\left( {{x_t}{\rm{;}}{{\overline a }_t}{x_0}{\rm{,}}\left( {1 - \overline a _t^2} \right){\rm{I}}} \right)}}\\
		\propto N\left( {{x_{t - 1}}{\rm{;}}\frac{{{a_t}\left( {1 - \overline a _{t - 1}^2} \right){x_t} + {{\overline a }_{t - 1}}\left( {1 - a_t^2} \right){x_0}}}{{1 - \overline a _t^2}}{\rm{,}}\frac{{\left( {1 - a_t^2} \right)\left( {1 - \overline a _t^2} \right)}}{{1 - \overline a _t^2}}{\rm{I}}} \right)
	\end{array}
\end{equation}

Let the mean of $q\left( {{x_{t - 1}}|{x_t},{x_0}} \right)$ be ${\mu _q} = \frac{{{a_t}\left( {1 - \overline a _{t - 1}^2} \right){x_t} + {{\overline a }_{t - 1}}\left( {1 - a_t^2} \right){x_0}}}{{1 - \overline a _t^2}}$, and its variance be $\sigma _q^2 = \frac{{\left( {1 - a_t^2} \right)\left( {1 - \overline a _{t - 1}^2} \right)}}{{1 - \overline a _t^2}}{\rm{I}}$. We first set the variance $\sigma _\theta ^{\rm{2}}$ of ${p_\theta }\left( {{x_{t - 1}}|{x_t}} \right)$ equal to the variance $\frac{{\left( {1 - a_t^2} \right)\left( {1 - \overline a _{t - 1}^2} \right)}}{{1 - \overline a _t^2}}{\rm{I}}$ of $q\left( {{x_{t - 1}}|{x_t},{x_0}} \right)$  , then minimize the Kullback-Leibler (KL) divergence between the two distributions to obtain:

\begin{equation}
	\label{deqn_ex1}
	\mathop {\arg \min }\limits_\theta  {D_{KL}}\left( {q\left( {{x_{t - 1}}|{x_t},{x_0}} \right)\left\| {{p_\theta }\left( {{x_{t - 1}}|{x_t}} \right)} \right.} \right) = \mathop {\arg \min }\limits_\theta  \frac{1}{{2\sigma _q^2\left( t \right)}}\left[ {\left\| {{\mu _\theta } - {\mu _q}} \right\|_2^2} \right]
\end{equation}

We then substitute ${x_0}$ in ${\mu _q} = \frac{{{a_t}\left( {1 - \overline a _{t - 1}^2} \right){x_t} + {{\overline a }_{t - 1}}\left( {1 - a_t^2} \right){x_0}}}{{1 - \overline a _t^2}}$ using the relational expression from the forward process:  ${x_t} = {\overline a _t}{x_0} + \sqrt {\left( {1 - \overline a _t^2} \right)} {\overline \varepsilon  _t},{\overline \varepsilon  _t} \sim N{\rm{(0,I)}}$ , resulting in:

\begin{equation}
	\label{deqn_ex1}
	{\mu _q} = \frac{1}{{{a_t}}}{x_t} - \frac{{\left( {1 - a_t^2} \right)}}{{\sqrt {\left( {1 - \overline a _t^2} \right)}  \cdot {a_t}}}{\overline \varepsilon  _t}
\end{equation}

Let ${\mu _\theta } = \frac{1}{{{a_t}}}{x_t} - \frac{{\left( {1 - a_t^2} \right)}}{{\sqrt {\left( {1 - \overline a _t^2} \right)}  \cdot {a_t}}}{f_\theta }\left( {{x_t},t} \right)$, then the final optimization objective can be transformed into:

\begin{equation}
	\label{deqn_ex1}
	\mathop {\arg \min }\limits_\theta  {D_{KL}}\left( {q\left( {{x_{t - 1}}|{x_t},{x_0}} \right)\left\| {{p_\theta }\left( {{x_{t - 1}}|{x_t}} \right)} \right.} \right) = \mathop {\arg \min }\limits_\theta  \frac{1}{{2\sigma _q^2\left( t \right)}}\frac{{{{\left( {1 - a_t^2} \right)}^2}}}{{\left( {1 - \overline a _t^2} \right) \cdot a_t^2}}\left[ {\left\| {{f_\theta }\left( {{x_t},t} \right) - {{\overline \varepsilon  }_t}} \right\|_2^2} \right]
\end{equation}

\subsection{Signal Attenuation Diffusion Model}
Based on ${x_t} = {a_t}{x_{t - 1}} + {b_t}{\varepsilon _t}{\rm{, }}{\varepsilon _t}\sim N{\rm{(0,I)}}$, our model introduces an attenuation factor ${k_t}$ , and the forward noise addition-attenuation process is expressed as:
\begin{equation}
	\label{deqn_ex1}
	{x_t} = {k_t}{\rm{(}}{a_t}{x_{t - 1}} + {b_t}{\varepsilon _t}{\rm{), }}{\varepsilon _t}\sim N{\rm{(0,I)}}
\end{equation}
Compared with the original model ${x_t} = {a_t}{x_{t - 1}} + {b_t}{\varepsilon _t}{\rm{, }}{\varepsilon _t}\sim N{\rm{(0,I)}} {\rm{,}}t = 1,2,...,T$ , our model adds the attenuation coefficient ${k_t}$ . Its purpose is to attenuate the signal while adding noise: when the original image evolves into pure noise, the signal value also tends to 0, and the image as a whole approximates a low-light image. This design integrates denoising and enhancement into a single task.
By expanding ${x_t} = {k_t}{\rm{(}}{a_t}{x_{t - 1}} + {b_t}{\varepsilon _t}{\rm{)}}$ step-by-step, we obtain:
\begin{equation}
	\label{deqn_ex1}
	\begin{array}{l}
		{x_t} = {k_t}{\rm{(}}{a_t}{x_{t - 1}} + {b_t}{\varepsilon _t}{\rm{)}}\\
		= {k_t}{a_t}{\rm{(}}{k_{t - 1}}{\rm{(}}{a_{t - 1}}{x_{t - 2}} + {b_{t - 1}}{\varepsilon _{t - 1}}{\rm{))}} + {k_t}{b_t}{\varepsilon _t}\\
		= {k_t}{k_{t - 1}}{a_t}{a_{t - 1}}{\rm{(}}{k_{t - 2}}{\rm{(}}{a_{t - 2}}{x_{t - 2}} + {b_{t - 2}}{\varepsilon _{t - 2}}{\rm{))}}\\
		+ {k_t}{k_{t - 1}}{a_t}{b_{t - 1}}{\varepsilon _{t - 1}} + {k_t}{b_t}{\varepsilon _t}\\
		\vdots \\
		= {k_t}{k_{t - 1}} \cdots {k_2}{k_1}{a_t}{a_{t - 1}} \cdots {a_2}{a_1}{x_0} + {k_t}{k_{t - 1}} \cdots {k_2}{k_1}{a_t}{a_{t - 1}}	\cdots {a_2}{b_1}{\varepsilon _1} +  \cdots  + {k_t}{k_{t - 1}}{a_t}{b_{t - 1}}{\varepsilon _{t - 1}} + {k_t}{b_t}{\varepsilon _t}
	\end{array}
\end{equation}
By superposing multiple independent Gaussian distributions ${\varepsilon _t}$ , we calculate the equivalent variance. Note that:
\begin{equation}
	\label{deqn_ex1}
	\begin{array}{l}
		\mathop \prod \limits_{s = 1}^t k_s^2\mathop \prod \limits_{s = 1}^t a_s^2 + \mathop \prod \limits_{s = 1}^t k_s^2\mathop \prod \limits_{s = 2}^t a_s^2b_1^2 +  \cdots  + \mathop \prod \limits_{s = t - 1}^t k_s^2\mathop \prod \limits_{s = t}^t a_s^2b_{t - 1}^2 + k_t^2b_t^2\\
		= \mathop \prod \limits_{s = 1}^t k_s^2{\rm{(}}\mathop \prod \limits_{s = 1}^t a_s^2 + \mathop \prod \limits_{s = 2}^t a_s^2b_1^2{\rm{)}} +  \cdots  + \mathop \prod \limits_{s = t - 1}^t k_s^2\mathop \prod \limits_{s = t}^t a_s^2b_{t - 1}^2 + k_t^2b_t^2\\
		= \mathop \prod \limits_{s = 1}^t k_s^2\mathop \prod \limits_{s = 2}^t a_s^2{\rm{(}}a_1^2 + b_1^2{\rm{) + }}\mathop \prod \limits_{s = 2}^t k_s^2\mathop \prod \limits_{s = 3}^t a_s^2b_2^2 +  \cdots  + \mathop \prod \limits_{s = t - 1}^t k_s^2\mathop \prod \limits_{s = t}^t a_s^2b_{t - 1}^2 + k_t^2b_t^2\\
		= \mathop \prod \limits_{s = 2}^t k_s^2\mathop \prod \limits_{s = 3}^t a_s^2{\rm{(}}k_{\rm{1}}^2a_{\rm{2}}^2{\rm{(}}a_1^2 + b_1^2{\rm{) + }}b_2^2{\rm{)}} +  \cdots  + \mathop \prod \limits_{s = t - 1}^t k_s^2\mathop \prod \limits_{s = t}^t a_s^2b_{t - 1}^2 + k_t^2b_t^2
	\end{array}
\end{equation}
Let $k_{t - 1}^4a_t^2 + b_t^2 = k_t^2,{\rm{and }}{k_0} = 1\left( {{\rm{i}}{\rm{.e}}{\rm{., }}a_1^2 + b_1^2 = k_1^2} \right)$, simplifying the above expression, we obtain:
\begin{equation}
	\label{deqn_ex1}
	\begin{array}{l}
		= \mathop \prod \limits_{s = 2}^t k_s^2\mathop \prod \limits_{s = 3}^t a_s^2{\rm{(}}k_{\rm{1}}^2a_{\rm{2}}^2{\rm{(}}a_1^2 + b_1^2{\rm{) + }}b_2^2{\rm{)}} +  \cdots  + \mathop \prod \limits_{s = t - 1}^t k_s^2\mathop \prod \limits_{s = t}^t a_s^2b_{t - 1}^2 + k_t^2b_t^2\\
		= \mathop \prod \limits_{s = 2}^t k_s^2\mathop \prod \limits_{s = 3}^t a_s^2{\rm{(}}k_{\rm{1}}^{\rm{4}}a_{\rm{2}}^2{\rm{ + }}b_2^2{\rm{)}} +  \cdots  + \mathop \prod \limits_{s = t - 1}^t k_s^2\mathop \prod \limits_{s = t}^t a_s^2b_{t - 1}^2 + k_t^2b_t^2\\
		{\rm{ = }}\mathop \prod \limits_{s = 2}^t k_s^2\mathop \prod \limits_{s = 3}^t a_s^2k_2^2 +  \cdots  + \mathop \prod \limits_{s = t - 1}^t k_s^2\mathop \prod \limits_{s = t}^t a_s^2b_{t - 1}^2 + k_t^2b_t^2\\
		= \mathop \prod \limits_{s = 3}^t k_s^2\mathop \prod \limits_{s = 4}^t a_s^2{\rm{(}}k_2^{\rm{4}}a_3^2{\rm{ + }}b_3^2{\rm{)}} +  \cdots  + \mathop \prod \limits_{s = t - 1}^t k_s^2\mathop \prod \limits_{s = t}^t a_s^2b_{t - 1}^2 + k_t^2b_t^2\\
		\vdots \\
		= \mathop \prod \limits_{s = t - 1}^t k_s^2\mathop \prod \limits_{s = t}^t a_s^2{\rm{(}}k_{t - 2}^2a_{t - 1}^2k_{t - 2}^2{\rm{ + }}b_{t - 1}^2{\rm{)}} + k_t^2b_t^2\\
		= k_t^2k_{t - 1}^2a_t^2k_{t - 1}^2 + k_t^2b_t^2\\
		= k_t^2{\rm{(}}k_{t - 1}^4a_t^2 + b_t^2{\rm{)}}\\
		{\rm{ = }}k_t^4
	\end{array}
\end{equation}
Thus, the equivalent variance of these independent Gaussian normal distributions is $k_t^4 - \mathop \prod \limits_{s = 1}^t k_s^2\mathop \prod \limits_{s = 1}^t a_s^2$ . Let $\mathop \prod \limits_{s = 1}^t k_s^{} = {\bar k_t},\mathop \prod \limits_{s = 1}^t a_s^{} = {\bar a_t}$ ; then we can derive the forward relational expressions:
\begin{equation}
	\label{deqn_ex1}
	\begin{array}{l}
		{x_t} = {k_t}{a_t}{x_{t - 1}} + {k_t}{b_t}{\varepsilon _t}\\
		{x_t} = {{\bar k}_t}{{\bar a}_t}{x_0} + \sqrt {k_t^4 - \bar k_t^2\bar a_t^2} {{\tilde \varepsilon }_t}\\
		{x_0} = \frac{1}{{{{\bar k}_t}{{\bar a}_t}}}{\rm{(}}{x_t} - \sqrt {k_t^4 - \bar k_t^2\bar a_t^2} {{\tilde \varepsilon }_t}{\rm{)}}
	\end{array}
\end{equation}
Where ${\tilde \varepsilon _t}$ is a standard Gaussian distribution distinct from ${\varepsilon _t}$ . From the Markov chain condition, we have:
\begin{equation}
	\label{deqn_ex1}
    \begin{array}{l}
	q{\rm{(}}{{\mathop{\rm x}\nolimits} _{t - 1}}{\rm{|}}{{\mathop{\rm x}\nolimits} _t}{\rm{) = }}\frac{{q{\rm{(}}{{\mathop{\rm x}\nolimits} _t}{\rm{|}}{{\mathop{\rm x}\nolimits} _{t - 1}}{\rm{)}}q{\rm{(}}{{\mathop{\rm x}\nolimits} _{t - 1}}{\rm{|}}{{\mathop{\rm x}\nolimits} _0}{\rm{)}}}}{{q{\rm{(}}{{\mathop{\rm x}\nolimits} _t}{\rm{|}}{{\mathop{\rm x}\nolimits} _0}{\rm{)}}}}\\
	= \frac{{N\left( {{x_t}{\rm{; }}{k_t}{a_t}{x_{t - 1}}{\rm{,(}}{k_t}{b_t}{{\rm{)}}^2}{\rm{I}}} \right)N\left( {{x_{t - 1}}{\rm{; }}{{\bar k}_{t - 1}}{{\bar a}_{t - 1}}{x_0}{\rm{,(}}k_{t - 1}^4 - \bar k_{t - 1}^2\bar a_{t - 1}^2{\rm{)I}}} \right)}}{{N\left( {{x_t}{\rm{; }}{{\bar k}_t}{{\bar a}_t}{x_0}{\rm{,(}}k_t^4 - \bar k_t^2\bar a_t^2{\rm{)I}}} \right)}}\\
	\propto \exp \left\{ { - \frac{1}{2}\left[ {\frac{{{{\left( {{x_t} - {k_t}{a_t}{x_{t - 1}}} \right)}^2}}}{{{{\left( {{k_t}{b_t}} \right)}^2}}} + \frac{{{{\left( {{x_{t - 1}} - {{\bar k}_{t - 1}}{{\bar a}_{t - 1}}{x_0}} \right)}^2}}}{{{{\left( {\sqrt {k_{t - 1}^4 - \bar k_{t - 1}^2\bar a_{t - 1}^2} } \right)}^2}}} - \frac{{{{\left( {{x_t} - {{\bar k}_t}{{\bar a}_t}{x_0}} \right)}^2}}}{{{{\left( {\sqrt {k_t^4 - \bar k_t^2\bar a_t^2} } \right)}^2}}}} \right]} \right\}\\
	\propto \exp \left\{ { - \frac{1}{2}\left[ {\frac{{ - 2{k_t}{a_t}{x_t}{x_{t - 1}} + k_t^2a_t^2x_{t - 1}^2}}{{k_t^2b_t^2}} + \frac{{x_{t - 1}^2 - 2{{\bar k}_{t - 1}}{{\bar a}_{t - 1}}{x_0}{x_{t - 1}}}}{{k_{t - 1}^4 - \bar k_{t - 1}^2\bar a_{t - 1}^2}} + C\left( {{x_t},{x_0}} \right)} \right]} \right\}\\
	\propto \exp \left\{ { - \frac{1}{2}\left[ {\left( {\frac{{k_t^2a_t^2}}{{k_t^2b_t^2}} + \frac{1}{{k_{t - 1}^4 - \bar k_{t - 1}^2\bar a_{t - 1}^2}}} \right)\left( {x_{t - 1}^2 - 2\frac{{\left( {\frac{{{a_t}{k_t}}}{{k_t^2b_t^2}}{x_t} + \frac{{{{\bar k}_{t - 1}}{{\bar a}_{t - 1}}}}{{k_{t - 1}^4 - \bar k_{t - 1}^2\bar a_{t - 1}^2}}{x_0}} \right)}}{{\frac{{k_t^2a_t^2}}{{k_t^2b_t^2}} + \frac{1}{{k_{t - 1}^4 - \bar k_{t - 1}^2\bar a_{t - 1}^2}}}}{x_{t - 1}}} \right) + C\left( {{x_t},{x_0}} \right)} \right]} \right\}\\
	\propto \exp \left\{ { - \frac{1}{2}\left[ {\left( {\frac{{k_t^2a_t^2\left( {k_{t - 1}^4 - \bar k_{t - 1}^2\bar a_{t - 1}^2} \right){\rm{ + }}k_t^2b_t^2}}{{k_t^2b_t^2\left( {k_{t - 1}^4 - \bar k_{t - 1}^2\bar a_{t - 1}^2} \right)}}} \right)\left( {x_{t - 1}^2 - 2\frac{{\left( {\frac{{{a_t}{k_t}}}{{k_t^2b_t^2}}{x_t} + \frac{{{{\bar k}_{t - 1}}{{\bar a}_{t - 1}}}}{{k_{t - 1}^4 - \bar k_{t - 1}^2\bar a_{t - 1}^2}}{x_0}} \right)}}{{\frac{{k_t^2a_t^2}}{{k_t^2b_t^2}} + \frac{1}{{k_{t - 1}^4 - \bar k_{t - 1}^2\bar a_{t - 1}^2}}}}{x_{t - 1}}} \right) + C\left( {{x_t},{x_0}} \right)} \right]} \right\}\\
	\propto \exp \left\{ { - \frac{1}{2}\left[ {\left( {\frac{{a_t^2k_{t - 1}^4 - a_t^2\bar k_{t - 1}^2\bar a_{t - 1}^2{\rm{ + }}b_t^2}}{{b_t^2\left( {k_{t - 1}^4 - \bar k_{t - 1}^2\bar a_{t - 1}^2} \right)}}} \right)\left( {x_{t - 1}^2 - 2\frac{{\left( {\frac{{{a_t}{k_t}}}{{k_t^2b_t^2}}{x_t} + \frac{{{{\bar k}_{t - 1}}{{\bar a}_{t - 1}}}}{{k_{t - 1}^4 - \bar k_{t - 1}^2\bar a_{t - 1}^2}}{x_0}} \right)}}{{\frac{{k_t^2a_t^2}}{{k_t^2b_t^2}} + \frac{1}{{k_{t - 1}^4 - \bar k_{t - 1}^2\bar a_{t - 1}^2}}}}{x_{t - 1}}} \right) + C\left( {{x_t},{x_0}} \right)} \right]} \right\}\\
	\propto \exp \left\{ { - \frac{1}{2}\left[ {\left( {\frac{1}{{\frac{{b_t^2\left( {k_{t - 1}^4 - \bar k_{t - 1}^2\bar a_{t - 1}^2} \right)}}{{k_t^2 - \bar k_{t - 1}^2\bar a_t^2}}}}} \right)\left( {x_{t - 1}^2 - 2\frac{{\left( {\frac{{{a_t}{k_t}}}{{k_t^2b_t^2}}{x_t} + \frac{{{{\bar k}_{t - 1}}{{\bar a}_{t - 1}}}}{{k_{t - 1}^4 - \bar k_{t - 1}^2\bar a_{t - 1}^2}}{x_0}} \right)b_t^2\left( {k_{t - 1}^4 - \bar k_{t - 1}^2\bar a_{t - 1}^2} \right)}}{{k_t^2 - \bar k_{t - 1}^2\bar a_t^2}}{x_{t - 1}}} \right) + C\left( {{x_t},{x_0}} \right)} \right]} \right\}\\
	\propto \exp \left\{ { - \frac{1}{2}\left[ {\left( {\frac{1}{{\frac{{b_t^2\left( {k_{t - 1}^4 - \bar k_{t - 1}^2\bar a_{t - 1}^2} \right)}}{{k_t^2 - \bar k_{t - 1}^2\bar a_t^2}}}}} \right)\left( {x_{t - 1}^2 - 2\left( {\frac{{{a_t}\left( {k_{t - 1}^4 - \bar k_{t - 1}^2\bar a_{t - 1}^2} \right)}}{{{k_t}\left( {k_t^2 - \bar k_{t - 1}^2\bar a_t^2} \right)}}{x_t} + \frac{{b_t^2{\rm{ }}{{\bar k}_{t - 1}}{{\bar a}_{t - 1}}}}{{\left( {k_t^2 - \bar k_{t - 1}^2\bar a_t^2} \right)}}{x_0}} \right){x_{t - 1}}} \right) + C\left( {{x_t},{x_0}} \right)} \right]} \right\}\\
	\propto N\left( {{x_{t - 1}}{\rm{; }}\left( {\frac{{{a_t}\left( {k_{t - 1}^4 - \bar k_{t - 1}^2\bar a_{t - 1}^2} \right)}}{{{k_t}\left( {k_t^2 - \bar k_{t - 1}^2\bar a_t^2} \right)}}{x_t} + \frac{{b_t^2{\rm{ }}{{\bar k}_{t - 1}}{{\bar a}_{t - 1}}}}{{\left( {k_t^2 - \bar k_{t - 1}^2\bar a_t^2} \right)}}{x_0}} \right){\rm{,}}\frac{{b_t^2\left( {k_{t - 1}^4 - \bar k_{t - 1}^2\bar a_{t - 1}^2} \right)}}{{k_t^2 - \bar k_{t - 1}^2\bar a_t^2}}{\rm{I}}} \right)
\end{array}
\end{equation}
i.e.,${\mu _q} = \frac{{{a_t}\left( {k_{t - 1}^4 - \bar k_{t - 1}^2\bar a_{t - 1}^2} \right)}}{{{k_t}\left( {k_t^2 - \bar k_{t - 1}^2\bar a_t^2} \right)}}{x_t} + \frac{{b_t^2{\rm{ }}{{\bar k}_{t - 1}}{{\bar a}_{t - 1}}}}{{\left( {k_t^2 - \bar k_{t - 1}^2\bar a_t^2} \right)}}{x_0}$ and $\sigma _q^2 = \frac{{b_t^2\left( {k_{t - 1}^4 - \bar k_{t - 1}^2\bar a_{t - 1}^2} \right)}}{{k_t^2 - \bar k_{t - 1}^2\bar a_t^2}}$.

Substituting ${x_0} = \frac{1}{{{{\bar k}_t}{{\bar a}_t}}}{\rm{(}}{x_t} - \sqrt {k_t^4 - \bar k_t^2\bar a_t^2} {\tilde \varepsilon _t}{\rm{)}}$ into ${\mu _q} = \frac{{{a_t}\left( {k_{t - 1}^4 - \bar k_{t - 1}^2\bar a_{t - 1}^2} \right)}}{{{k_t}\left( {k_t^2 - \bar k_{t - 1}^2\bar a_t^2} \right)}}{x_t} + \frac{{b_t^2{\rm{ }}{{\bar k}_{t - 1}}{{\bar a}_{t - 1}}}}{{\left( {k_t^2 - \bar k_{t - 1}^2\bar a_t^2} \right)}}{x_0}$, we obtain:
\begin{equation}
	\label{deqn_ex1}
	\begin{array}{l}
		{\mu _q} = \frac{{{a_t}\left( {k_{t - 1}^4 - \bar k_{t - 1}^2\bar a_{t - 1}^2} \right)}}{{{k_t}\left( {k_t^2 - \bar k_{t - 1}^2\bar a_t^2} \right)}}{x_t} + \frac{{b_t^2{\rm{ }}{{\bar k}_{t - 1}}{{\bar a}_{t - 1}}}}{{\left( {k_t^2 - \bar k_{t - 1}^2\bar a_t^2} \right)}}\frac{1}{{{{\bar k}_t}{{\bar a}_t}}}{\rm{(}}{x_t} - \sqrt {k_t^4 - {{\bar k}_t}^2{{\bar a}_t}^2} {{\tilde \varepsilon }_t}{\rm{)}}\\
		{\rm{ = }}\left( {\frac{{{a_t}\left( {k_{t - 1}^4 - \bar k_{t - 1}^2\bar a_{t - 1}^2} \right)}}{{{k_t}\left( {k_t^2 - \bar k_{t - 1}^2\bar a_t^2} \right)}} + \frac{{b_t^2{\rm{ }}{{\bar k}_{t - 1}}{{\bar a}_{t - 1}}}}{{\left( {k_t^2 - \bar k_{t - 1}^2\bar a_t^2} \right)}}\frac{1}{{{{\bar k}_t}{{\bar a}_t}}}} \right){x_t} - \frac{{b_t^2{\rm{ }}{{\bar k}_{t - 1}}{{\bar a}_{t - 1}}}}{{\left( {k_t^2 - \bar k_{t - 1}^2\bar a_t^2} \right)}}\frac{1}{{{{\bar k}_t}{{\bar a}_t}}}\sqrt {k_t^4 - \bar k_t^2\bar a_t^2} {{\tilde \varepsilon }_t}\\
		= \left( {\frac{{{a_t}\left( {k_{t - 1}^4 - \bar k_{t - 1}^2\bar a_{t - 1}^2} \right)}}{{{k_t}\left( {k_t^2 - \bar k_{t - 1}^2{{\bar a}_t}^2} \right)}} + \frac{{b_t^2{\rm{ }}}}{{\left( {k_t^2 - \bar k_{t - 1}^2\bar a_t^2} \right)}}\frac{1}{{{k_t}{a_t}}}} \right){x_t} - \frac{{b_t^2{\rm{ }}}}{{\left( {k_t^2 - \bar k_{t - 1}^2\bar a_t^2} \right)}}\frac{1}{{{k_t}{a_t}}}{k_t}\sqrt {k_t^2 - \bar k_t^2\bar a_t^2} {{\tilde \varepsilon }_t}\\
		= \frac{{a_t^2k_{t - 1}^4 - a_t^2\bar k_{t - 1}^2\bar a_{t - 1}^2 + b_t^2}}{{{k_t}{a_t}\left( {k_t^4 - \bar k_{t - 1}^2\bar a_t^2} \right)}}{x_t} - \frac{{b_t^2{\rm{ }}}}{{{a_t}\sqrt {k_t^2 - \bar k_{t - 1}^2\bar a_t^2} }}{{\tilde \varepsilon }_t}\\
		= \frac{{k_t^4 - \bar k_{t - 1}^2\bar a_t^2}}{{{k_t}{a_t}\left( {k_t^4 - \bar k_{t - 1}^2\bar a_t^2} \right)}}{x_t} - \frac{{b_t^2{\rm{ }}}}{{{a_t}\sqrt {k_t^2 - \bar k_{t - 1}^2\bar a_t^2} }}{{\tilde \varepsilon }_t}\\
		= \frac{1}{{{k_t}{a_t}}}{x_t} - \frac{{b_t^2{\rm{ }}}}{{{a_t}\sqrt {k_t^2 - \bar k_{t - 1}^2\bar a_t^2} }}{{\tilde \varepsilon }_t}
	\end{array}
\end{equation}
Let $\sigma _\theta ^2 = \sigma _q^2$ , and let ${\mu _\theta } = \frac{1}{{{k_t}{a_t}}}{x_t} - \frac{{b_t^2}}{{\sqrt {\left( {\frac{{k_t^2}}{{a_t^2}} - \overline k _{t - 1}^2\overline a _{t - 1}^2} \right)}  \cdot a_t^2}}{f_\theta }\left( {{x_t},t} \right)$ . Then the final optimization objective can be transformed into:
\begin{equation}
	\label{deqn_ex1}
	\begin{array}{l}
		\mathop {\arg \min }\limits_\theta  {D_{KL}}\left( {q\left( {{x_{t - 1}}|{x_t},{x_0}} \right)\left\| {{p_\theta }\left( {{x_{t - 1}}|{x_t}} \right)} \right.} \right)
		= \mathop {\arg \min }\limits_\theta  \frac{1}{{2\sigma _q^2\left( t \right)}}\frac{{{{\left( {b_t^2} \right)}^2}}}{{\left( {\frac{{k_t^2}}{{a_t^2}} - \overline k _{t - 1}^2\overline a _{t - 1}^2} \right) \cdot a_t^4}}\left[ {\left\| {{f_\theta }\left( {{x_t},t} \right) - {{\tilde \varepsilon }_t}} \right\|_2^2} \right]
	\end{array}
\end{equation}

\subsection{DDIM Sampling}
Step-by-step iterative computation using DDPM typically requires substantial computational resources. To address this, this study combines the multi-scale pyramid and Denoising Diffusion Implicit Models (DDIM) sampling to improve the efficiency of image processing. By reducing image resolution, multi-scale pyramid sampling can effectively decrease the computational load. In contrast, DDIM optimizes the sampling logic within the diffusion process, significantly enhancing inference speed without compromising generation quality. The integration of these two techniques enables the construction of an efficient and high-fidelity generation framework.

Through mathematical derivation, DDIM breaks the constraints of the traditional Markov chain. It eliminates the need for retraining the DDPM and only requires modifications to the sampler. The modified sampler thus achieves a significant acceleration in sampling speed while preserving the quality of generated images.

Assume $p\left( {{x_{t - 1}}|{x_t},{x_0}} \right)$  satisfies ${p_\theta }\left( {{x_{t - 1}}|{x_t},{x_0}} \right) = N\left( {{x_{t - 1}}{\rm{;}}m{x_0} + n{x_t},{\sigma ^2}} \right)$  (this differs from the Markov chain process of DDPM). From the forward noise addition formula, we know: ${x_t} = {\bar k_t}{\bar a_t}{x_0} + \sqrt {k_t^4 - \bar k_t^2\bar a_t^2} {\tilde \varepsilon _t}$ . Substituting this into ${x_{t - 1}} = m{x_0} + n{x_t} + \sigma \varepsilon $ , we get:
\begin{equation}
	\label{deqn_ex1}
	{x_{t - 1}} = m{x_0} + n{\bar k_t}{\bar a_t}{x_0} + n\sqrt {k_t^4 - \bar k_t^2\bar a_t^2} {\tilde \varepsilon _t} + \sigma \varepsilon 
\end{equation}
where ${\tilde \varepsilon _t}$ , ${\tilde \varepsilon _{t - 1}}$ , and ${\varepsilon}$ are mutually independent and all follow the standard Gaussian distribution. Similarly, the forward noise addition formula can be used to derive: ${x_{t - 1}} = {\bar k_{t - 1}}{\bar a_{t - 1}}{x_0} + \sqrt {k_{t - 1}^4 - \bar k_{t - 1}^2\bar a_{t - 1}^2} {\tilde \varepsilon _{t - 1}}$ . By combining like terms, we require the corresponding coefficients to match, leading to:
\begin{equation}
	\label{deqn_ex1}
	\begin{array}{*{20}{c}}
		{m + n{{\bar k}_t}{{\bar a}_t} = {{\bar k}_{t - 1}}{{\bar a}_{t - 1}}}
		{{n^2}(k_t^4 - \bar k_t^2\bar a_t^2) + {\sigma ^2} = k_{t - 1}^4 - \bar k_{t - 1}^2\bar a_{t - 1}^2}
	\end{array}
\end{equation}
Solving the above equations, we obtain:
\begin{equation}
	\label{deqn_ex1}
	\begin{array}{l}
		m = {{\bar k}_{t - 1}}{{\bar a}_{t - 1}} - {{\bar k}_t}{{\bar a}_t}\frac{{\sqrt {k_{t - 1}^4 - \bar k_{t - 1}^2\bar a_{t - 1}^2 - {\sigma ^2}} }}{{\sqrt {k_t^4 - \bar k_t^2\bar a_t^2} }}
		n = \frac{{\sqrt {k_{t - 1}^4 - \bar k_{t - 1}^2\bar a_{t - 1}^2 - {\sigma ^2}} }}{{\sqrt {k_t^4 - \bar k_t^2\bar a_t^2} }}
	\end{array}
\end{equation}
Substituting back into ${x_{t - 1}} = m{x_0} + n{x_t} + \sigma \varepsilon $ , we obtain:
\begin{equation}
	\label{deqn_ex1}
	\begin{array}{l}
		{x_{t - 1}} = ({{\bar k}_{t - 1}}{{\bar a}_{t - 1}} - {{\bar k}_t}{{\bar a}_t}\frac{{\sqrt {k_{t - 1}^4 - \bar k_{t - 1}^2\bar a_{t - 1}^2 - {\sigma ^2}} }}{{\sqrt {k_t^4 - \bar k_t^2\bar a_t^2} }}){x_0}
		+ (\frac{{\sqrt {k_{t - 1}^4 - \bar k_{t - 1}^2\bar a_{t - 1}^2 - {\sigma ^2}} }}{{\sqrt {k_t^4 - \bar k_t^2\bar a_t^2} }}){x_t} + \sigma \varepsilon 
	\end{array}
\end{equation}
Using the same principle as reverse denoising, we can substitute ${x_0}$  in the above formula:
\begin{equation}
	\label{deqn_ex1}
	\begin{array}{l}
		{x_{t - 1}} = ({{\bar k}_{t - 1}}{{\bar a}_{t - 1}} - {{\bar k}_t}{{\bar a}_t}\frac{{\sqrt {k_{t - 1}^4 - \bar k_{t - 1}^2\bar a_{t - 1}^2 - {\sigma ^2}} }}{{\sqrt {k_t^4 - \bar k_t^2\bar a_t^2} }})\frac{1}{{{{\bar k}_t}{{\bar a}_t}}}{\rm{(}}{x_t} - \sqrt {k_t^4 - \bar k_t^2\bar a_t^2} {{\tilde \varepsilon }_t}{\rm{)}}
		+ (\frac{{\sqrt {k_{t - 1}^4 - \bar k_{t - 1}^2\bar a_{t - 1}^2 - {\sigma ^2}} }}{{\sqrt {k_t^4 - \bar k_t^2\bar a_t^2} }}){x_t} + \sigma \varepsilon 
	\end{array}
\end{equation}
The final form derived is:
\begin{equation}
	\label{deqn_ex1}
	{x_{t - 1}} = \frac{{{{\bar k}_{t - 1}}{{\bar a}_{t - 1}}}}{{{{\bar k}_t}{{\bar a}_t}}}({x_t} - \sqrt {k_t^4 - \bar k_t^2\bar a_t^2} {\tilde \varepsilon _t}) + \sqrt {k_{t - 1}^4 - \bar k_{t - 1}^2\bar a_{t - 1}^2 - {\sigma ^2}} {\tilde \varepsilon _t} + \sigma \varepsilon
\end{equation}
Throughout the derivation, the Markov chain process is not used. Thus, ${{t - 1}}$  in the above formula can be any previous step ${p}$ :
\begin{equation}
	\label{deqn_ex1}
	{x_p} = \frac{{{{\bar k}_p}{{\bar a}_p}}}{{{{\bar k}_t}{{\bar a}_t}}}({x_t} - \sqrt {k_t^4 - \bar k_t^2\bar a_t^2} {\tilde \varepsilon _t}) + \sqrt {k_p^4 - \bar k_p^2\bar a_p^2 - {\sigma ^2}} {\tilde \varepsilon _t} + \sigma \varepsilon 
\end{equation}

\subsection{Noise Scheduling Optimization of SADM}

\subsubsection{Recurrence Constraint Design and Motivation}

The proposed recurrence constraint $k_{t-1}^4 a_t^2 + b_t^2 = k_t^2$ is not an arbitrary formulation, but a necessary result derived from second-moment (signal energy) scheduling tailored to the physical nature of low-light image degradation, while ensuring mathematical consistency for the customized diffusion process of SADM.

In diffusion models, the second-moment of a random vector $\boldsymbol{x}$, denoted as $\mathbb{E}\|\boldsymbol{x}\|^2$, quantifies the total signal energy as the expectation of the sum of squared pixel values, serving as a core metric for designing noise scheduling strategies. For the standard DDPM, the canonical constraint $a_t^2 + b_t^2 = 1$ is built on the principle of energy conservation: given the normalized state $\mathbb{E}\|\boldsymbol{x}_{t-1}\|^2 \approx 1$ and standard Gaussian noise with $\mathbb{E}\|\varepsilon_t\|^2 = 1$ ($\varepsilon_t \sim \mathcal{N}(0, \mathbf{I})$), its forward process $\boldsymbol{x}_t = a_t \boldsymbol{x}_{t-1} + b_t \varepsilon_t$ maintains a constant total energy $\mathbb{E}\|\boldsymbol{x}_t\|^2 = 1$. This design is well-suited for general image generation tasks but mismatched with low-light enhancement, as the intrinsic degradation of low-light scenes is characterized by continuous signal energy decay rather than constant energy.

To physically simulate the degradation of normal-light images into extreme low-light ones, we build on the $[0,1]$ pixel range and introduce a global attenuation coefficient $k_t = 0.999 k_{t-1}$ with $k_0=1$ to enforce monotonic signal energy decay. The core goal of our second-moment scheduling is thus defined as: for the normalized initial normal-light energy $\mathbb{E}\|\boldsymbol{x}_0\|^2 \approx 1$, the expected signal energy at step $t$ follows $\mathbb{E}\|\boldsymbol{x}_t\|^2 = k_t^4$. This fourth-power trajectory stems from the multiplicative attenuation in SADM's forward process $\boldsymbol{x}_t = k_t (a_t \boldsymbol{x}_{t-1} + b_t \varepsilon_t)$: the factor $k_t$ is squared when computing the second-moment, and inductive coupling with the prior step's energy $k_{t-1}^4$ yields the fourth-power form, which precisely models the nonlinear noise amplification effect in low-light imaging where weak signals suffer more severe attenuation.

We derive this recurrence constraint by computing the second-moment of $\boldsymbol{x}_t$ in the forward process, leveraging the statistical independence between $\varepsilon_t$ and $\boldsymbol{x}_{t-1}$:
\[
\mathbb{E}\|\boldsymbol{x}_t\|^2 = \mathbb{E}\left\|k_t\left(a_t \boldsymbol{x}_{t-1} + b_t \varepsilon_t\right)\right\|^2 = k_t^2 \left( a_t^2 \mathbb{E}\|\boldsymbol{x}_{t-1}\|^2 + b_t^2 \mathbb{E}\|\varepsilon_t\|^2 \right)
\]
Substituting the inductive hypothesis $\mathbb{E}\|\boldsymbol{x}_{t-1}\|^2 = k_{t-1}^4$, the standard Gaussian noise property $\mathbb{E}\|\varepsilon_t\|^2 = 1$, and our second-moment scheduling target $\mathbb{E}\|\boldsymbol{x}_t\|^2 = k_t^4$, we obtain:
\[
k_t^4 = k_t^2 \left( a_t^2 k_{t-1}^4 + b_t^2 \right)
\]
Dividing both sides by $k_t^2$ (valid for all $t$ as $k_t > 0$) naturally yields the recurrence constraint $k_{t-1}^4 a_t^2 + b_t^2 = k_t^2$.

\begin{figure*}[htbp]
	\vskip 0.2in
	\begin{center}
		\vspace*{-15pt} 
		\centering
		\includegraphics[width=\linewidth]{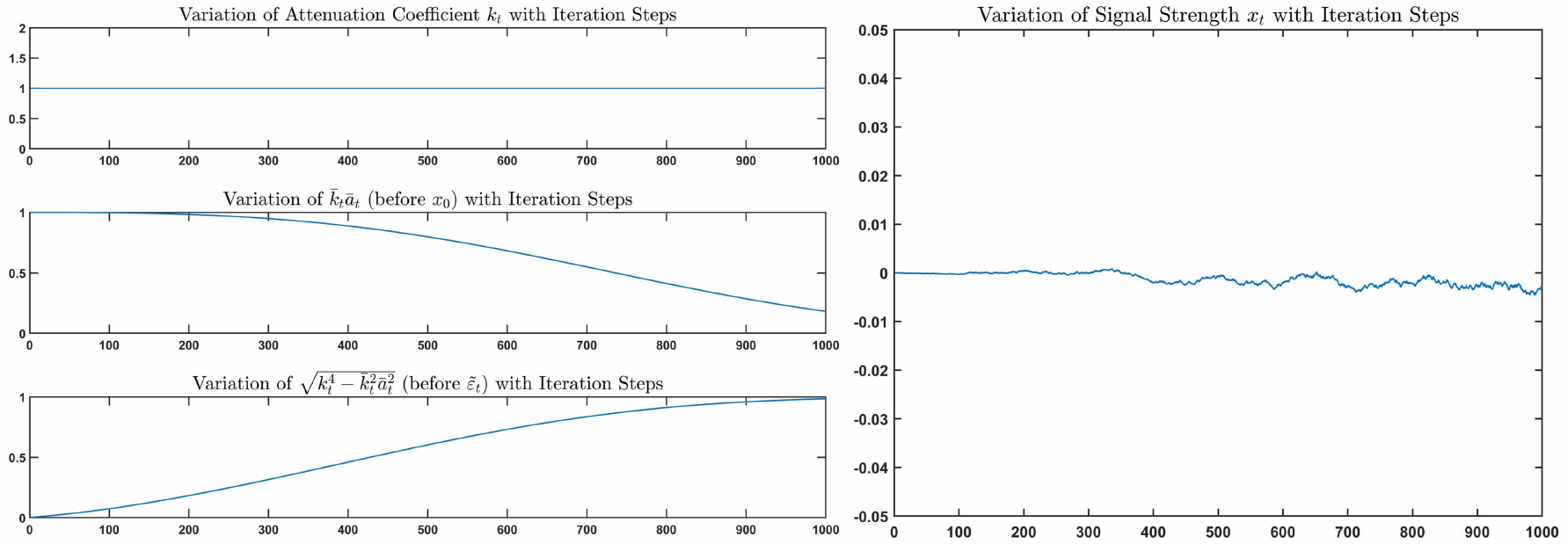}
		\caption{Variation of DDPM Noise Scheduling Coefficients with Iteration Steps(value range of images is set to [-1, 1]).}
		\label{fig:k1}
	\end{center}
\end{figure*}

\begin{figure*}[htbp]
	\vskip 0.2in
	\begin{center}
		\centering
		\includegraphics[width=\linewidth]{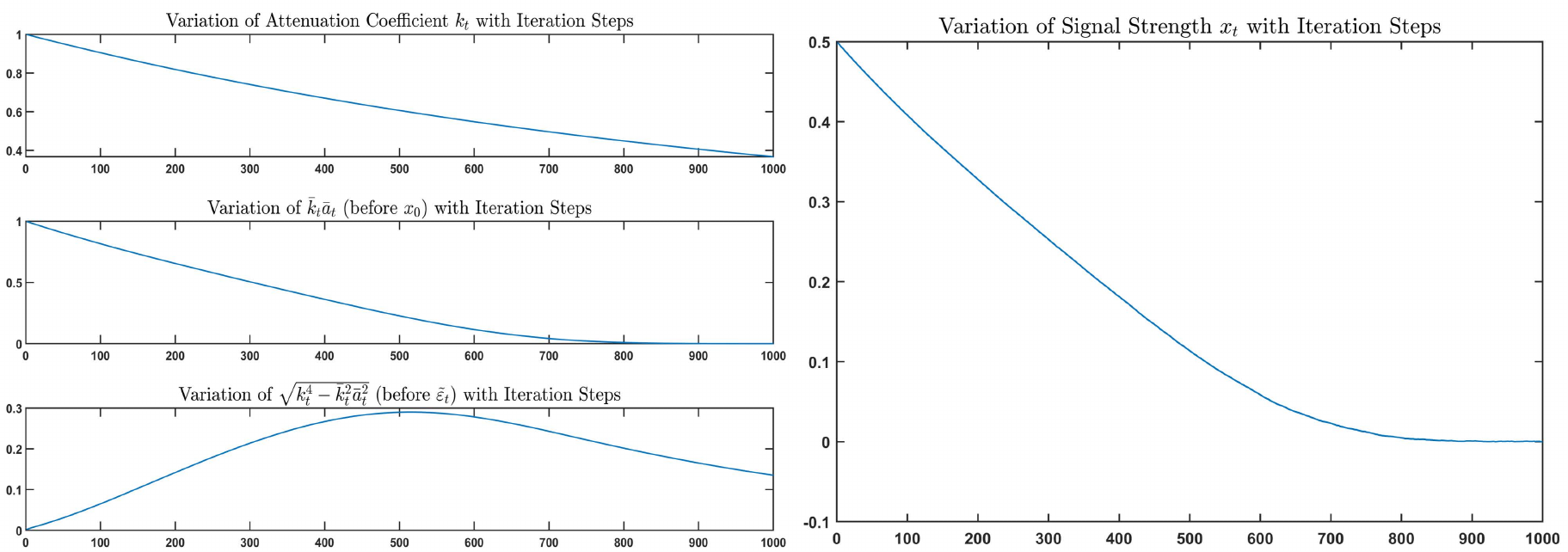}
		\caption{Variation of SADM Noise Scheduling Coefficients with Iteration Steps(value range of images is set to [0, 1]).}
		\label{fig:k999}
	\end{center}
			\vspace*{-10pt} 
\end{figure*}

This constraint maintains strict theoretical compatibility with the standard DDPM: when setting $k_t \equiv 1$ (i.e., no signal energy decay), our constraint directly degenerates to the classic DDPM constraint $a_t^2 + b_t^2 = 1$, confirming that SADM's diffusion framework is a natural generalized extension of the standard DDPM to energy-attenuating low-light degradation scenarios. Numerically, we validate that SADM's forward process strictly adheres to the second-moment schedule $\mathbb{E}\|\boldsymbol{x}_t\|^2 \approx k_t^4$, guaranteeing consistency between theoretical design and practical implementation. As illustrated in Figs. \ref{fig:k1} and \ref{fig:k999}, this design enables a steady decay of signal energy from approximately 0.5 (normal-light level) to near zero (extreme low-light level) over 1000 diffusion steps, while avoiding numerical instability from overly fast attenuation (e.g., $k_t=0.99$ in Fig. \ref{fig:k99}) and insufficient degradation from overly slow attenuation (e.g., $k_t=0.9999$ in Fig. \ref{fig:k9999}). By aligning the forward diffusion process with the intrinsic physical nature of low-light image degradation, this constraint eliminates the need for additional correction networks—an ad-hoc component in traditional DDPM-based low-light methods that often introduces color bias and visual artifacts.

\subsubsection{Noise Scheduling Design and Optimization for SADM}

The noise added by diffusion models follows a standard normal distribution. To conform to the characteristics of this distribution, the pixel value range of images is typically set to [-1, 1]. However, the pixel values of input low-light images are mostly negative, so attenuating these values to 0 essentially amounts to enhancing the mean value (shifting from negative values to 0). Therefore, the pixel value range of images is set to [0, 1] in our model.

When standard normal noise is applied with the data range constrained to [0, 1], some pixel values may exceed this range. Nevertheless, the favorable mathematical properties of standard normal noise simplify model implementation and ensure the stability and consistency of training. Meanwhile, it can effectively enhance image contrast and detail representation. This noise injection strategy has been proven effective in practice. Even if clipping operations are required to handle out-of-range pixel values, the overall performance of the model is still satisfactory.

In the original DDPM model ${x_t} = {a_t}{x_{t - 1}} + {b_t}{\varepsilon _t}{\rm{, }}{\varepsilon _t}\sim N{\rm{(0,I)}}$, the value range of  $b_t^{\rm{2}}$ is set to 0.00004–0.01 and $a_t^2 = 1 - b_t^2$. As the number of iteration steps increases, the coefficient preceding ${x_t}$ decreases, while the coefficient preceding ${\varepsilon _t}$ increases. This process characterizes the gradual transformation of a signal into disordered noise. In our model ${x_t} = {k_t}{\rm{(}}{a_t}{x_{t - 1}} + {b_t}{\varepsilon _t}{\rm{), }}{\varepsilon _t}\sim N{\rm{(0,I)}}$, the decay coefficient ${k_t} = 0.999 \times {k_{t - 1}},t > 1\left( {{k_0} = 1} \right)$ is adopted, and the value range of $b_t^{\rm{2}}$ remains 0.00004–0.01. The coefficient ${a_t}$ is calculated by $k_{t - 1}^4a_t^2 + b_t^2 = k_t^2,and {k_0} = 1,{\rm{i}}{\rm{.e}}{\rm{.,}}a_1^2 + b_1^2 = k_1^2$, yielding the following attenuation law of ${x_t} = {\bar k_t}{\bar a_t}{x_0} + \sqrt {k_t^4 - {{\bar k}_t}^2{{\bar a}_t}^2} {\hat \varepsilon _t}$. Specifically, the coefficient preceding ${x_t}$ also exhibits a decreasing trend, whereas the coefficient preceding  ${\varepsilon _t}$ first increases and then decreases. Overall, the intensity of the signal still shows a decaying tendency.

As illustrated in Fig. \ref{fig:k1} and Fig. \ref{fig:k999}, we analyze the behavior of noise scheduling coefficients under different pixel value ranges. The original DDPM sets the pixel value range of images to $[-1, 1]$. For normal-light input images, the signal strength centers around 0 on average. Under this prior, the variation of signal strength with noise scheduling, which is depicted in the right panel of Fig. \ref{fig:k1}, exhibits only slight fluctuations near 0 and shows no decaying trend.

\begin{figure*}[htbp]
	\vskip 0.2in
	\begin{center}
		\vspace*{-10pt} 
		\centering
		\includegraphics[width=\linewidth]{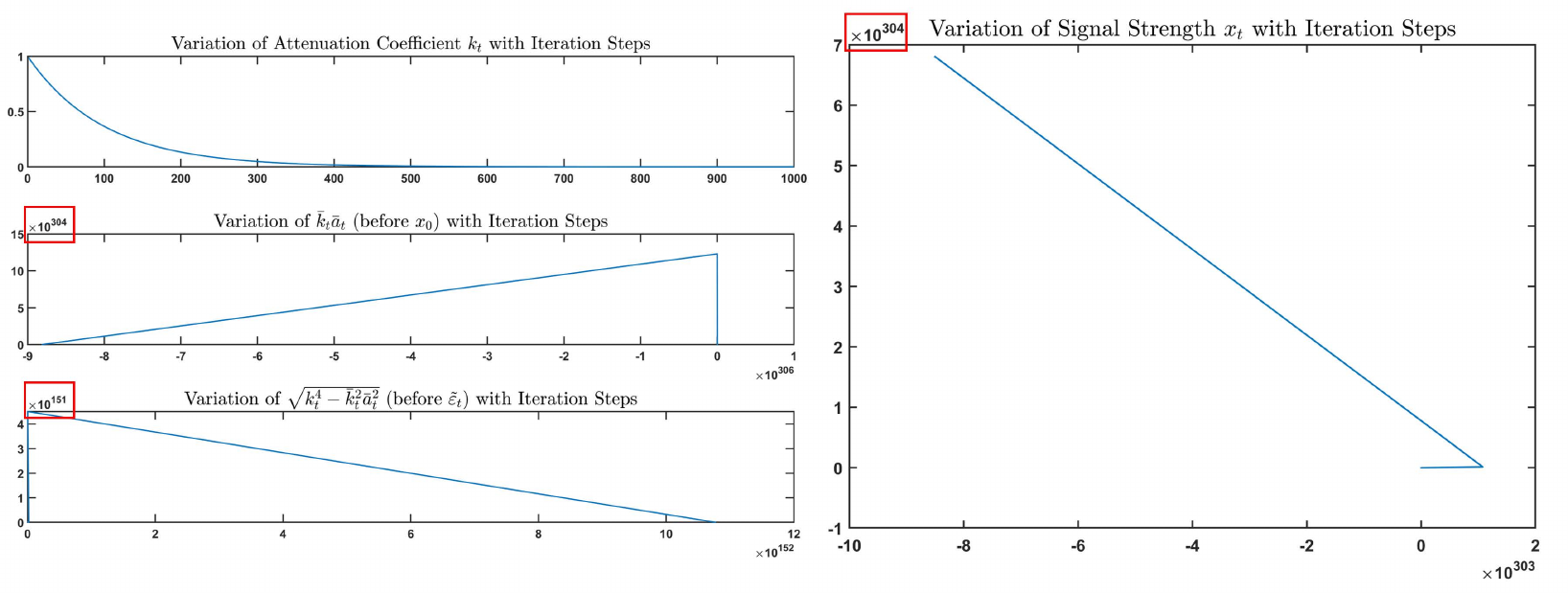}
		\caption{Variation of SADM Noise Scheduling Coefficients with 0.99.}
		\label{fig:k99}
	\end{center}
\end{figure*}

\begin{figure*}[htbp]
	\vskip 0.2in
	\begin{center}
		\vspace*{-10pt} 
		\centering
		\includegraphics[width=\linewidth]{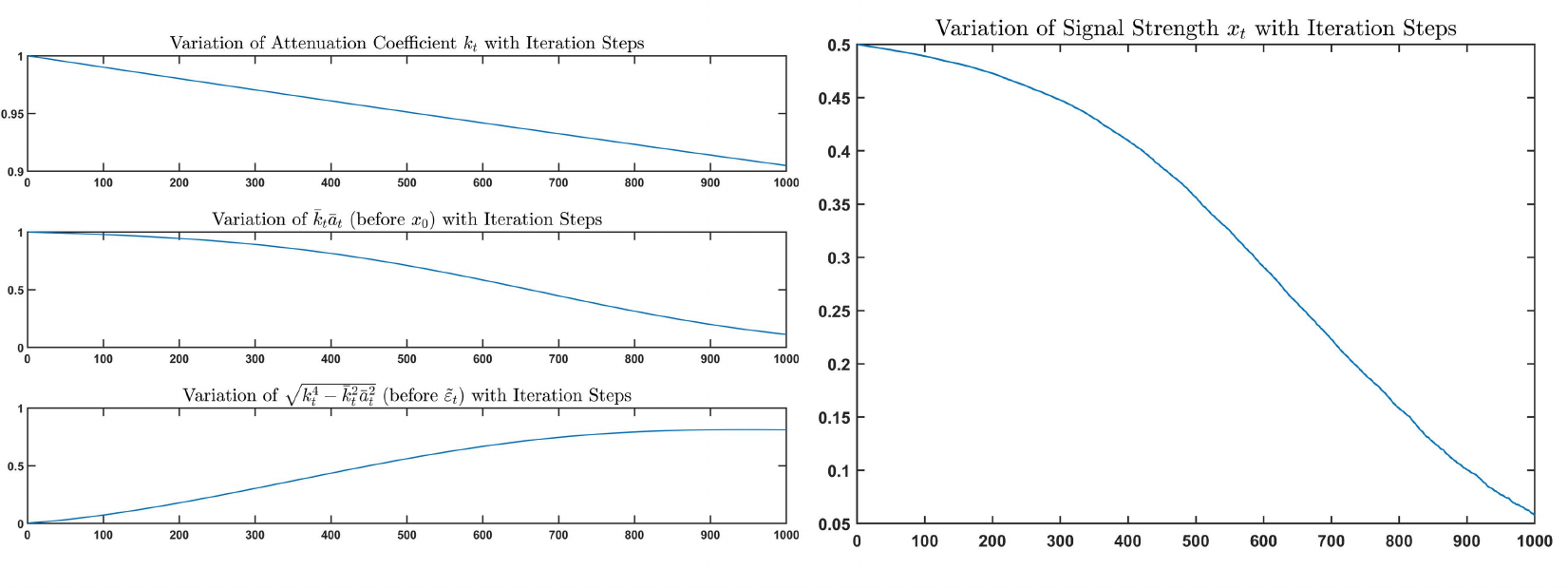}
		\caption{Variation of SADM Noise Scheduling Coefficients with 0.9999.}
		\label{fig:k9999}
	\end{center}
\end{figure*}

In contrast, our SADM model constrains the pixel value range to $[0, 1]$. Under this range, normal-light images exhibit an average signal strength around 0.5. To effectively simulate the gradual degradation to extremely low illumination within a finite number of sampling steps, we adopt a noise scheduling coefficient of $k_t = 0.999k_{t-1}$. This value represents a practical balance. A smaller coefficient, such as 0.99(Fig.\ref{fig:k99}), could introduce numerical instability during sampling, whereas a larger one like 0.9999(Fig.\ref{fig:k9999}) would lead to an insufficient decay rate, preventing the signal strength from adequately approaching zero within the typical span of 1000 steps. Consequently, under this scheduling, the signal strength exhibits a gradual decay as the number of sampling steps increases, eventually approaching 0.

This difference in the signal starting point underpins a fundamental divergence in the physical processes the two models emulate. In DDPM-like models, the forward process starts from a normal-light image, which is in an ordered state characterized by an average signal strength around 0, and adds noise until it reaches pure Gaussian noise, a disordered state with zero signal strength. The reverse sampling process then starts from this zero-signal noise and aims to recover a normal-light image. Therefore, the entire process essentially models the transition between a normal-light state and noise, a pathway which has weak inherent relevance to the characteristics of low-light images. Many algorithms introduce additional correction networks precisely to strengthen this tenuous connection to low-light conditions, a strategy that can introduce theoretical limitations and artifacts such as color bias.

Our SADM framework operates within the $[0, 1]$ pixel range, thereby aligning the starting point of its reverse process with the low signal strength characteristic of low-light images, where signal strength denotes the normalized pixel intensity. Consequently, the forward process models the degradation of a normal-light image into a state that resembles low-light noise, characterized by disorder and decaying signal. The reverse process then captures the recovery from this disordered, near-zero-signal state back to an ordered normal-light image with enhanced signal. This intrinsic alignment establishes a direct relevance of the entire process to low-light enhancement. It therefore obviates the need for a dedicated correction network to enforce this connection, which in turn avoids the associated theoretical and practical limitations, such as color bias.


\subsection{Prior Condition with Dehazing for SADM}
This subsection details the dehazing-based preprocessing module of the SADM model, a critical prior step designed to enhance the initial visibility of low-light images before they are fed into the diffusion model with the aim of simplifying the enhancement pipeline, uncovering fine-grained dark-region features, and alleviating color distortion.

The core rationale of this preprocessing lies in treating the inverted low-light image as a pseudo-hazy image. By applying dehazing to this pseudo-hazy image and inverting the dehazed result, we effectively recover obscured details in dark regions, mitigate color distortion inherent to low-light imaging, and ultimately boost the efficacy of the input prior for subsequent diffusion-based enhancement.

Given the original low-light image with its grayscale matrix denoted as $X_{L}$, we first invert $X_{L}$ to generate the pseudo-hazy image, whose grayscale matrix is defined as $X_{{L}_d} = 1 - X_{L}$. To align with the input requirements of the subsequent dehazing module, all grayscale values of $X_{L}$ and $X_{{L}_d}$ are normalized to the range $[0, 1]$.

The dehazing of $X_{{L}_d}$ adheres to the classic Atmospheric Scattering Model, which depends on two key parameters including atmospheric light intensity $A$ and modified medium transmittance $T_r$. The baseline dehazing formulation reads  ${X_{{E_d}}} = \frac{{{X_{{L_d}}} - A}}{{{T_r}}} + A$. To suppress noise in low-light scenarios and avoid over-dehazing caused by overly small transmittance values, we introduce an adaptive bias term to compute $T_r$:
\begin{equation}
	\label{deqn_ex1}
	{T_r} = {t_0} + \omega  \cdot (1 - \omega  \cdot mi{n_{c \in \{ r,g,b\} }}(X_{{L_d}}^c))
\end{equation}
Here ${t_0} = mean({X_L})$ (the global grayscale mean of the original low-light image), $\omega  = 1 - {t_0}$ serves as the fidelity coefficient, and we set $A = 1$ as the optimal value for normalized low-light images. Substituting $A=1$ and $T_r$ into the baseline model yields the dehazed pseudo-hazy image, expressed as:
\begin{equation}
	\label{deqn_ex1}
	{X_{{E_d}}} = \frac{{{X_{{L_d}}} - 1}}{{{t_0} + \omega  \cdot (1 - \omega  \cdot mi{n_{c \in \{ r,g,b\} }}(X_{{L_d}}^c))}} + 1
\end{equation}

To simplify the computational pipeline while preserving the benefits of dark-region feature exposure and color distortion reduction, we derive the preprocessed enhanced image $X_E = 1 - X_{E_d}$ via algebraic simplification of the above formulation. The result is:
\begin{equation}
	\label{deqn_ex1}
	X_E = 1 - X_{E_d} = \frac{1 - X_{L_d}}{t_0 + \omega \cdot \left(1 - \omega \cdot \min_{c \in \{r,g,b\}}(X_{L_d}^c)\right)} = \frac{X_L}{t_0 + \omega \cdot \left(1 - \omega \cdot \min_{c \in \{r,g,b\}}(X_{L_d}^c)\right)}
\end{equation}

We adopt a stage-specific strategy for post-processing  $X_{E}$ to balance model performance and inference efficiency. During training, we further convert $X_{E}$ to the YCbCr color space and perform gamma correction on the luminance channel. This operation strengthens the model’s ability to capture subtle dark-region features by enhancing the contrast of under-exposed areas. In contrast, during test-time sampling, we omit the gamma correction step to prioritize inference speed and deployment practicality. The simplified preprocessing pipeline can still maintain satisfactory enhancement quality without gamma correction, making it more suitable for real-time or near-real-time application scenarios.

This simplified formulation eliminates redundant inversion/dehazing/re-inversion steps in the original pipeline, enabling direct computation of $X_{E}$ from the raw low-light image $X_{L}$. Critically, the simplification retains the core advantages of the dehazing-based prior by uncovering subtle features in under-exposed dark regions, mitigating color distortion caused by low-light sensor noise, and streamlining the overall enhancement workflow for integration with the diffusion model.
\begin{figure}[htbp]
	\vskip 0.2in
	\begin{center}
		\centering
		\vspace*{-10pt} 
		\includegraphics[width=\linewidth]{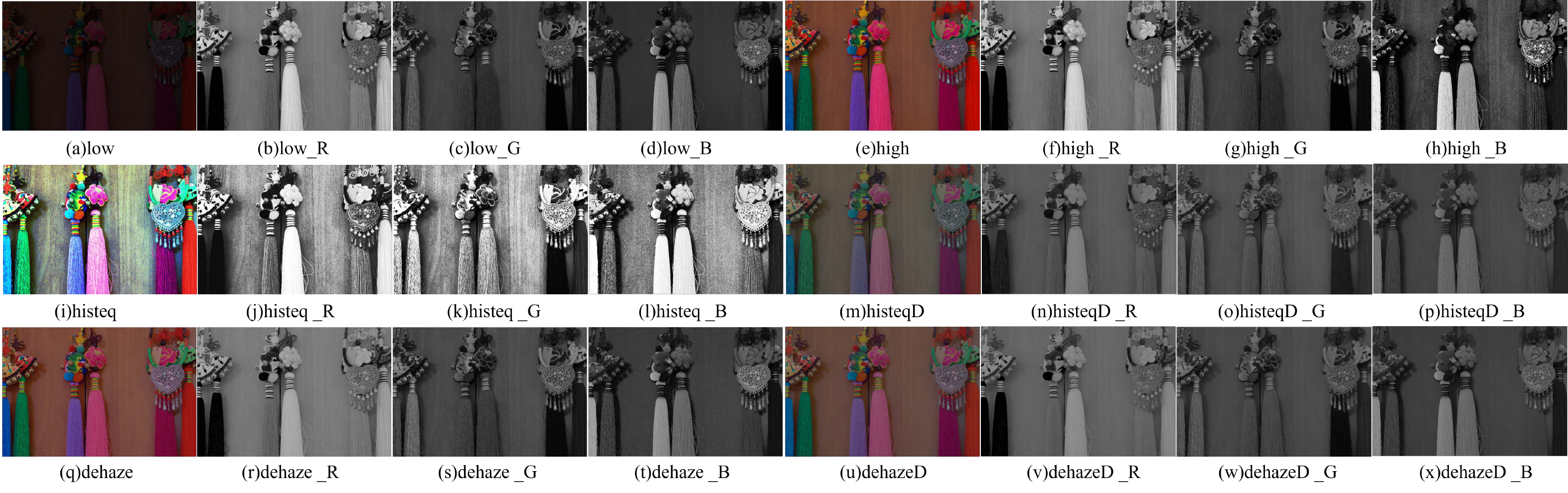}
		\caption{Visual Comparison of Dehazing and Histogram Equalization Preprocessing With RGB Channel Distribution Analysis for Diffusion Based Low Light Image Enhancement.
		}
		\label{fig:haze}
	\end{center}
\end{figure}

As shown in Fig.\ref{fig:haze}, the superiority of the simplified dehazing preprocessing pipeline over histogram equalization is visually validated through comprehensive RGB channel distribution analysis and qualitative comparison of enhancement results. The figure presents the original low-light input image, its corresponding high-light reference image, as well as the outputs generated by the two preprocessing methods and their subsequent diffusion-based enhancement results. 
Notably, the histogram equalization method redistributes pixel intensities independently across the R, G, and B channels, which disrupts the inherent channel differences that are clearly observable in the high-light reference. This independent channel redistribution leads to noticeable color deviation in the enhanced outputs. 
In contrast, the simplified dehazing preprocessing preserves the original RGB channel differences while enhancing the visibility of dark-region details such as fine textures and intricate patterns. When integrated with the diffusion model, the dehazing-based preprocessing provides a high-fidelity prior that enables the model to produce outputs with more accurate color reproduction and clearer dark-region features, which are closely aligned with the high-light reference. The RGB channel decomposition results further illustrate the distinct behavior of the two preprocessing strategies, with histogram equalization reducing channel-level differences and dehazing maintaining them to ensure color fidelity.

\section{Detailed Experiments}
\subsection{Training Details}
We detail the key training configuration of SADM for reproducibility, with full implementation details and environment configurations to be released in our open-source code upon paper acceptance.

We use three standard low-light enhancement datasets: LOLv1 (485 train/15 test pairs), LOLv2-Real (689/100), and LOLv2-Synthetic (900/100). During training, images are randomly cropped to patches of 160×160 pixels. For validation, images are padded to sizes divisible by 32. Input preprocessing involves concatenating the low-light image with 10-channel positional encodings and a histogram-equalized version of the image (applied randomly during training). We employ data shuffling for stability but do not use flipping or brightness augmentation. For evaluating generalization on unpaired data, we initialize the model with weights pre-trained on the LOLv1 dataset.

The model is implemented with PyTorch 1.13.1 (CUDA 11.7) and trained on a single NVIDIA GeForce RTX 4090 D GPU (24.5 GB VRAM). We fix the random seed (manual\_seed=1), enable CUDA prefetching with pinned memory, and use a batch size of 8 with 4 data-loading workers.

Our core denoising network is an SR3UNet. It takes a 13-channel input (3 channels low light image + 4 positional channels + 3 dehazed image + 3 channels high light image) and outputs a 3-channel image. The model uses 64 base channels, with channel multipliers [1, 2, 4, 8, 8] across five resolution stages, defining a U-Net with downsampling to 1/16 of the input size. It employs Group Normalization (32 groups), self-attention layers at the 16×16 resolution, and 2 residual blocks per resolution level with a dropout rate of 0.2. This network is integrated into a GaussianDiffusion model configured for conditional image generation at 128×128 resolution, using a specific pyramid noise schedule [1,1,1,2,2,2,4,4,4,4]. The total number of parameters is 97.8 million.

We train the model for 1,000,000 iterations using the Adam optimizer. The learning rate is set to 1e-4 for the generator (denoising network) and 0.002 for the discriminator and any auxiliary components. The learning rate is decayed by a factor of 0.5 at iterations 50k, 75k, 100k, 150k, and 200k using a MultiStepLR scheduler. The diffusion process uses a linear noise schedule over 1000 timesteps, with $b_t^{\rm{2}}$
increasing from $4\times 10^{-5}$ to $10^{-2}$.
At test time, we use the DDIM sampler with ddim\_eta=0.0, and accelerate sampling to 10 timesteps. A full training run takes approximately 3 days to complete.

\subsection{Supplementary Experimental Results}
Here we will illustrate more experimental results.
\begin{figure}[htbp]
	\vskip 0.2in
	\begin{center}
		\centering
		\vspace*{-10pt} 
		\includegraphics[width=\linewidth]{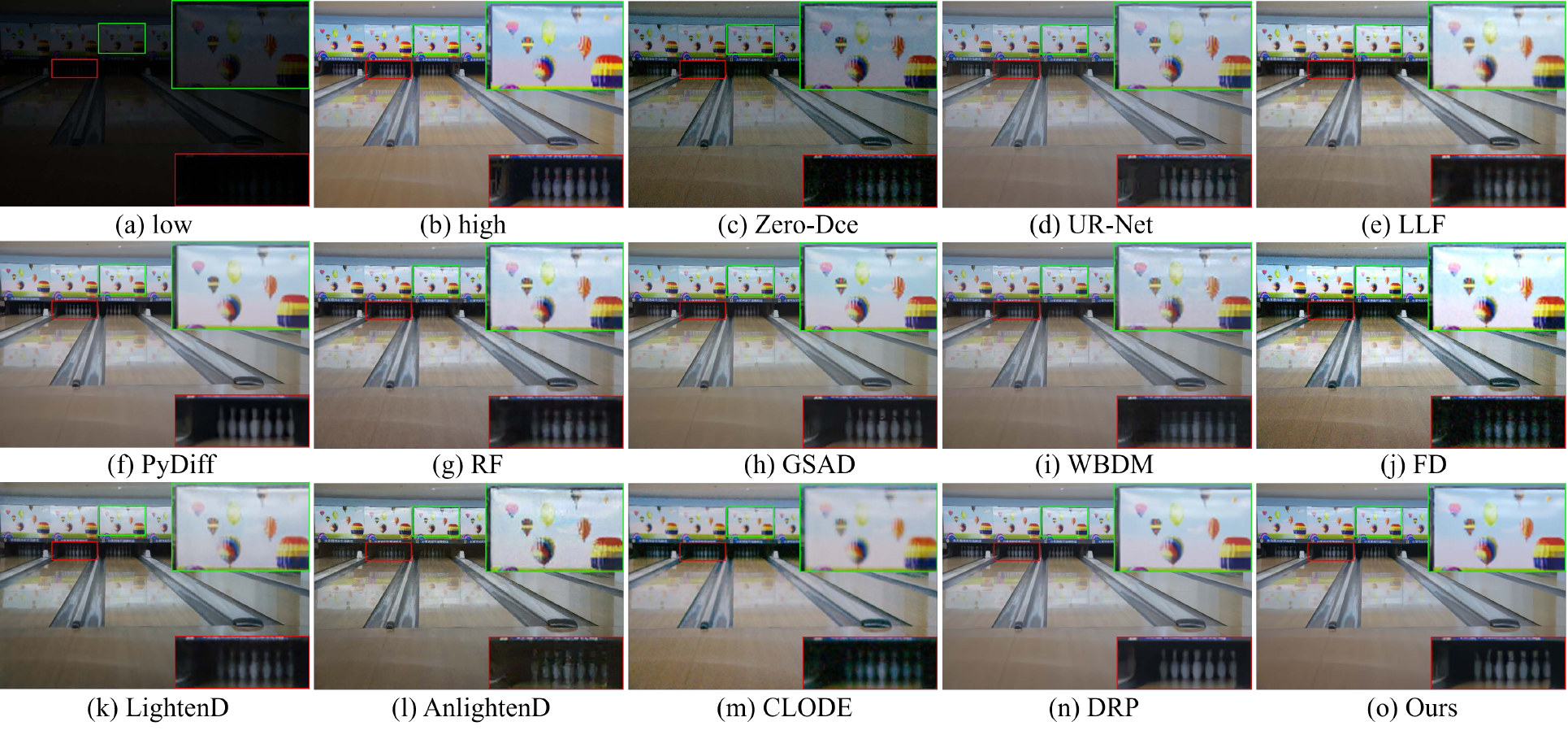}
		\caption{	Visual effect comparison of different algorithms on the LOLv1test dataset:(a)low; (b)high; (c)Zero-Dce; (d)UR-Net; (e)LLF; (f)PyDiff; (g)RF; (h)GSAD; (i)WBDM; (j)FD; (k)LightenD; (l)AnlightenD; (m)CLODE; (m)DRP; (n)Ours. Local magnified views are presented in the red and green boxes.	}
		\label{fig:669}
	\end{center}
\end{figure}

\begin{figure*}[htbp]
	\vskip 0.2in
	\begin{center}
		\vspace*{-20pt} 
		\centering
		\includegraphics[width=\linewidth]{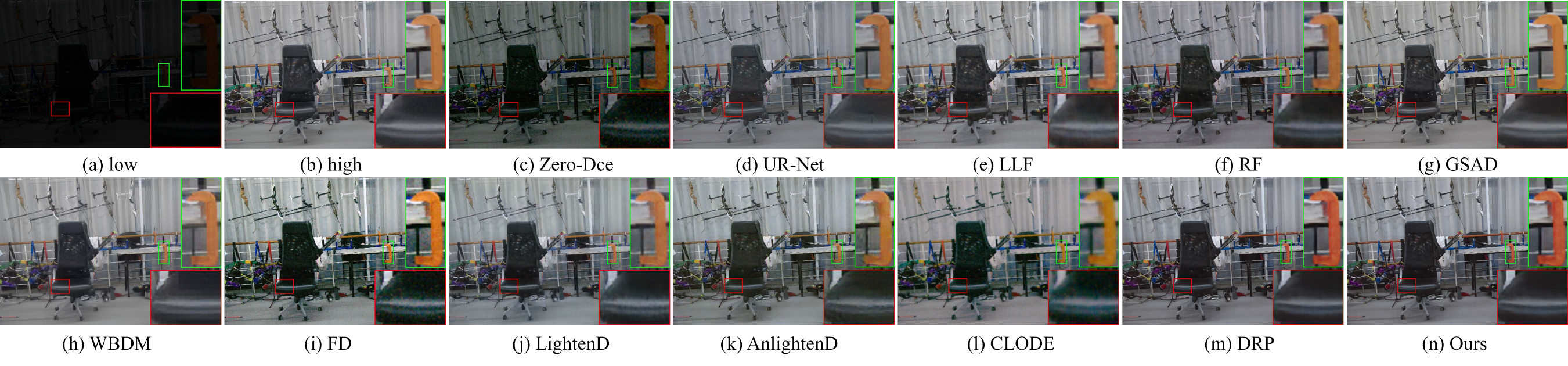}
		\caption{Visual effect comparison of different algorithms on the LOLv2\_real dataset:(a)low; (b)high; (c)Zero-Dce; (d)UR-Net; (e)LLF; (f)RF; (g)GSAD; (h)WBDM; (i)FD; (j)LightenD; (k)AnlightenD; (l)CLODE; (m)DRP; (n)Ours. Local magnified views are presented in the red and green boxes.}
		\label{fig:5_v2}
	\end{center}
			\vspace*{-10pt} 
\end{figure*}
\begin{figure*}[htbp]
	\vskip 0.2in
	\begin{center}
		\vspace*{-10pt} 
		\centering
		\includegraphics[width=\linewidth]{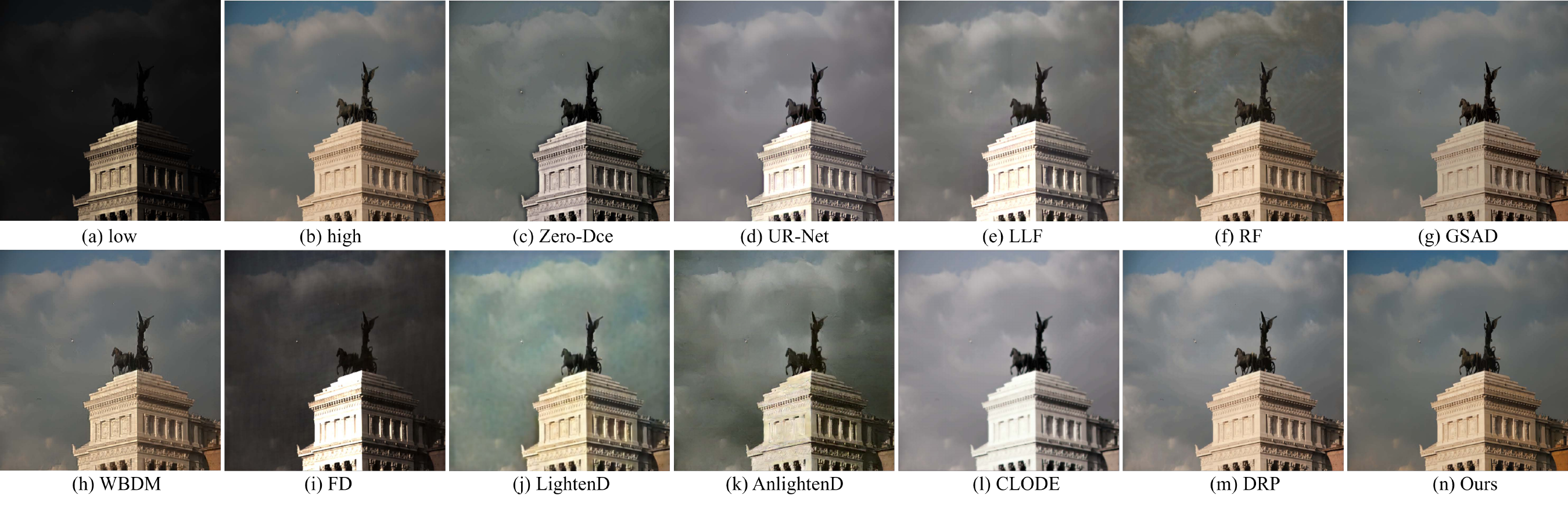}
		\caption{Visual effect comparison of different algorithms on the LOLv2\_syn dataset:(a)low; (b)high; (c)Zero-Dce; (d)UR-Net; (e)LLF; (f)RF; (g)GSAD; (h)WBDM; (i)FD; (j)LightenD; (k)AnlightenD; (l)CLODE; (m)DRP; (n)Ours.}
		\label{fig:85_v2syn}
	\end{center}
\end{figure*}

\begin{figure*}[htbp]
	\vskip 0.2in
	\begin{center}
		\vspace*{-20pt} 
		\centering
		\includegraphics[width=\linewidth]{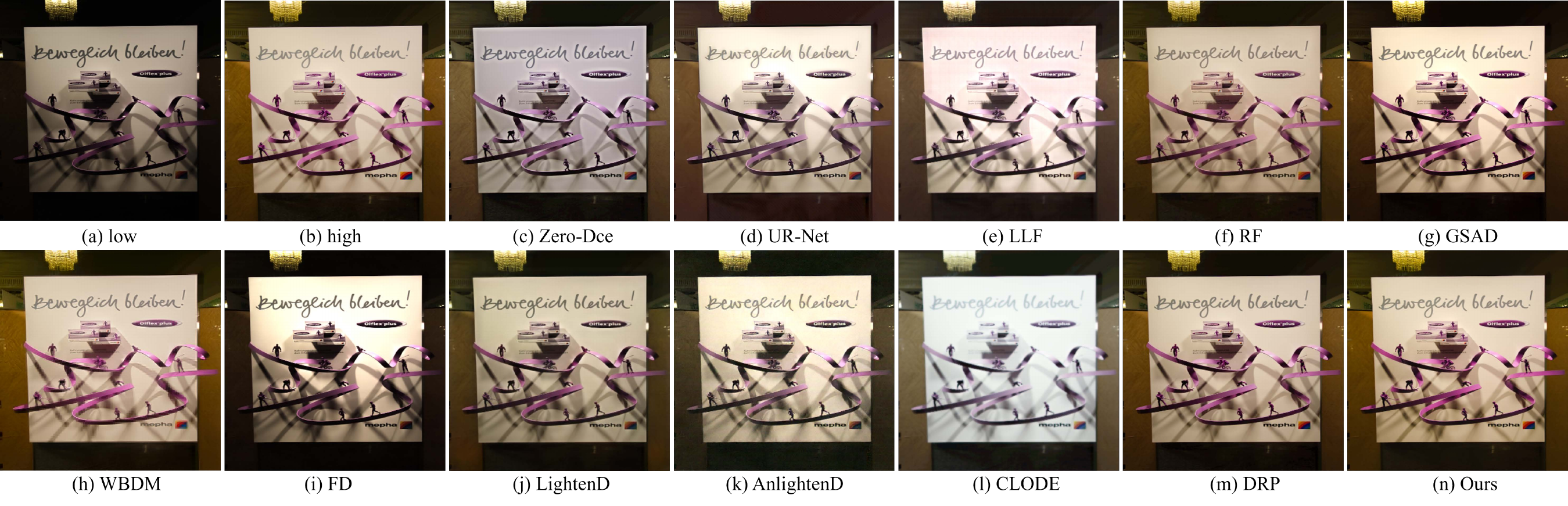}
		\caption{Visual effect comparison of different algorithms on the LOLv2\_syn dataset.}
		\label{fig:99_v2syn}
	\end{center}
\end{figure*}
\begin{figure*}[htbp]
	\vskip 0.2in
	\begin{center}
		\centering
		\vspace*{-25pt} 
		\includegraphics[width=\linewidth]{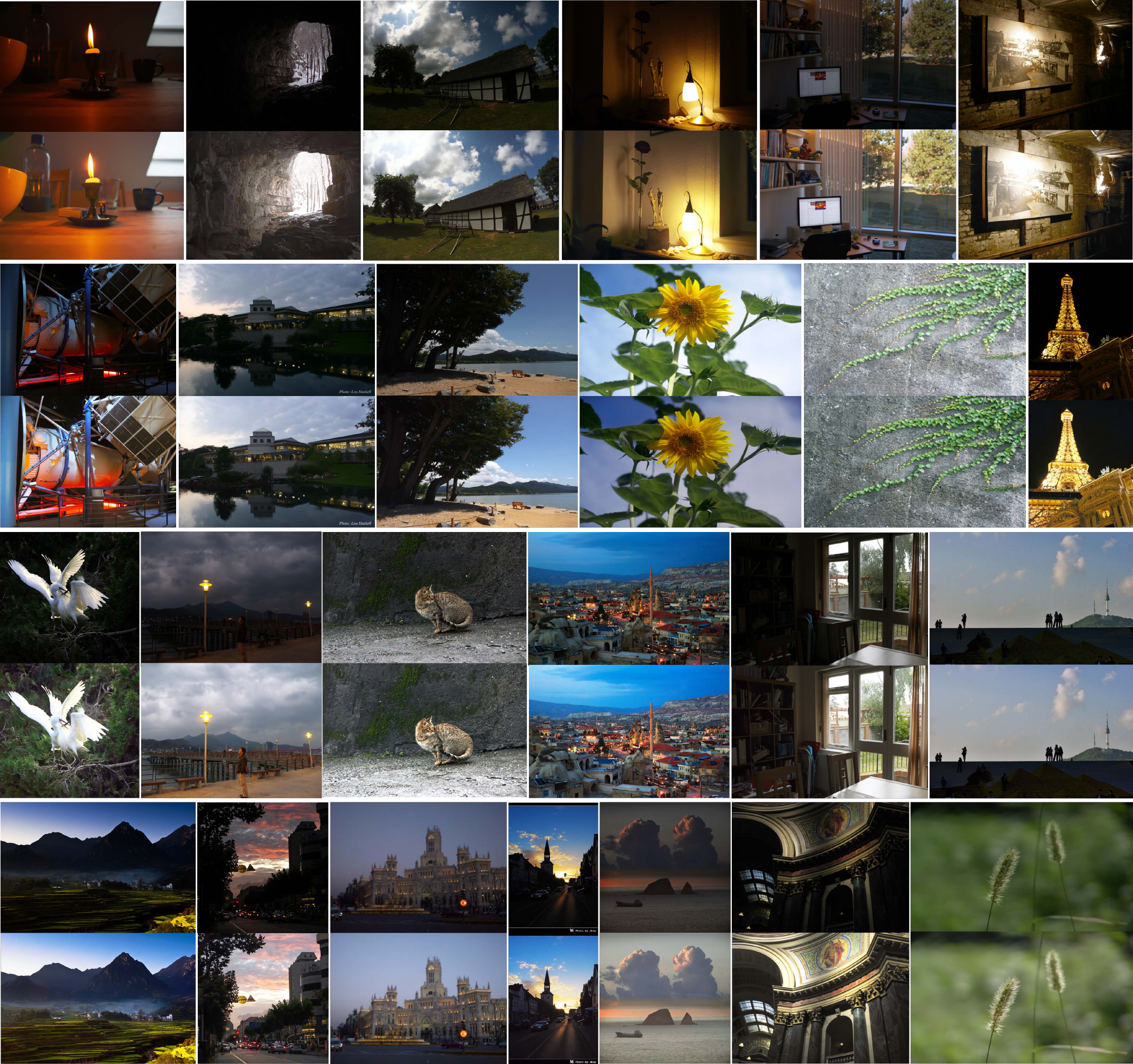}
		\caption{SADM's visual results on the unpaired dataset. }
		\label{fig:Ab2}
		\vspace*{-15pt} 
	\end{center}
\end{figure*}

\end{document}